\documentclass[11pt]{article}

\usepackage[final]{acl}

\usepackage{times}
\usepackage{latexsym}
\usepackage{authblk}
\usepackage{hyperref}
\usepackage{CJKutf8}
\usepackage{times}
\usepackage{graphicx}
\usepackage{xcolor}
\usepackage{float}
\usepackage{amsmath}
\usepackage{booktabs} 
\usepackage{multirow} 
\usepackage{multicol} 
\usepackage{subfig}
\usepackage{CJKutf8}
\usepackage{tabularx}
\usepackage{algorithm}
\usepackage{algpseudocode}
\usepackage{amsmath}
\usepackage[T1]{fontenc}

\usepackage[utf8]{inputenc}

\usepackage{microtype}

\usepackage{inconsolata}

\usepackage{graphicx}

\title{MAG: A Web-Agent Benchmark and Harness for Multimodal Action and Guide Generation}

\author{
\textbf{Chengguang Gan}\textsuperscript{1}, \textbf{Hanjun Wei}\textsuperscript{2}, \textbf{Yunhao Liang}\textsuperscript{2}, \\
\textbf{Zhixi Cai}\textsuperscript{3}, \textbf{Qinghao Zhang}\textsuperscript{4}, \textbf{Shiwen Ni}\textsuperscript{5} \\
\small\textsuperscript{1}Independent Researcher, \quad \textsuperscript{2}University of Chinese Academy of Sciences, \\
\small\textsuperscript{3}Monash University, \quad \textsuperscript{4}Pusan National University, \\
\small\textsuperscript{5}Shenzhen University of Advanced Technology \\
\small
\textbf{Correspondence:} \href{mailto:chengguangg1024@gmail.com}{chengguangg1024@gmail.com}
}

\begin{document}
\pagestyle{plain} 
\maketitle

\begin{abstract}
Digital Adoption Platforms (DAPs) are embedded overlays widely used on web
systems to guide users through operations inside a page, helping them get
started with unfamiliar interfaces quickly. Completing a real task, however,
rarely means clicking a few buttons on a single page: it takes a sequence of
actions that unfolds across changing page states. Prior studies have also
treated automated web agent actions and guide text generation as two separate
problems, and most of them feed models textual page representations such as
the DOM or accessibility trees rather than the rendered screens that humans
actually operate on. In this work we introduce \textbf{MAG}, the first
benchmark that unifies task execution and guide writing into a single
\textbf{M}ultimodal \textbf{A}ction and \textbf{G}uide task, with two
grounding schemes over screenshots: Set-of-Mark element selection and raw
pixel coordinates. We further build a complete harness for this compound
task, covering annotation with LLM assistance and human verification,
training, evaluation in live environments, and joint metrics for actions and
guides. With this harness we evaluate frontier API models and open multimodal
models, and report detailed analyses. Finally, we design a GRPO training
method augmented with expert trajectories, which nearly doubles the success
rate of a supervised 9B agent (from 6.9\% to 13.2\%) and improves guide
quality at the same time. Even the strongest model completes fewer than
40\% of the tasks, leaving ample room for future research.
\end{abstract}

\section{Introduction}
\label{sec:intro}

\begin{figure}[t!]
\centering
\includegraphics[width=\columnwidth]{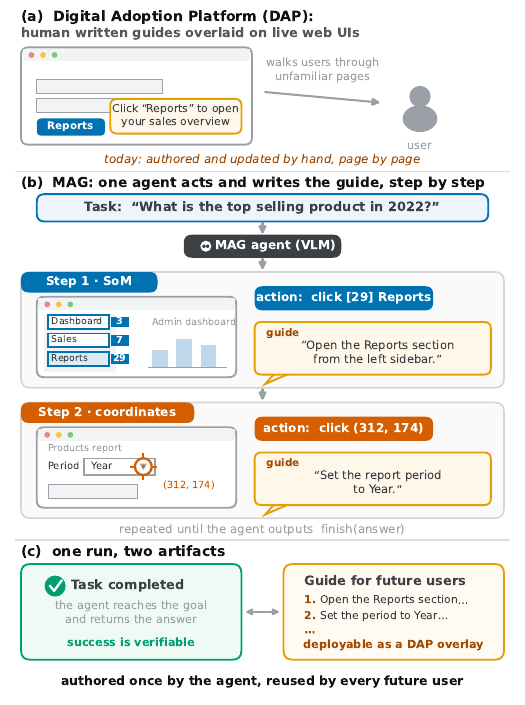}
\caption{Overview of MAG. (a) A Digital Adoption Platform overlays human
written guides on live web pages; today these guides are authored and
updated by hand. (b) The MAG task asks a single agent to complete the task
and to write a guide at every step, under two grounding schemes: Set-of-Mark
element selection (Step 1) and raw pixel coordinates (Step 2). (c) One run
yields two artifacts: a verifiable task outcome and a guide that future
users can reuse as a DAP overlay.}
\label{fig:teaser}
\end{figure}

Commercial Digital Adoption Platforms (DAPs) overlay guidance on web systems
to help new users operate complex, unfamiliar interfaces. When a user faces an unfamiliar page, the platform highlights the
element that matters and shows a short instruction next to it
(Figure~\ref{fig:teaser}(a)). This convenience rests on manual labor: vendors locate the target element
of every step and write its guide text, and every redesign forces both to
be redone. Because one goal often spans several pages and a chain of
dependent operations, guidance for a whole system is expensive to author and
maintain.

Web agents are a natural way to remove this labor, but the two relevant
research lines have developed separately. Agent benchmarks score task
completion alone~\cite{zhou2024webarena,deng2023mind2web,yao2022webshop,koh2024visualwebarena}:
none asks the agent to produce, or scores, the instruction a human would
need at each step. Guide generation has been studied in the opposite direction, producing guidance for one given page~\cite{gan2026guideweb} without executing anything, on textual page representations, although multimodal agents
show that acting from rendered screens is
practical~\cite{zheng2024gpt,he2024webvoyager,hong2024cogagent}.

We address this gap with MAG, a benchmark for Multimodal Action and Guide
generation. To our knowledge, it is the first benchmark in which an agent
must both complete a multistep task on a live website and write, at every
step, the guide sentence a future user would need at that point
(Figure~\ref{fig:teaser}(b)). The agent observes the site through screenshots and acts under one of two
grounding schemes: Set-of-Mark selection~\cite{yang2023set} over numbered
elements, or raw pixel coordinates. MAG
builds on the six live websites of WebArena~\cite{zhou2024webarena}: 581
tasks with verified success demonstrations, split into 407 training and 174
test. For 563 of them the demonstrations are reannotated into gold
trajectories, 4,760 Set-of-Mark and 5,779 coordinate steps, each step
carrying a guide drafted by an LLM and then corrected by human annotators.

We release the full harness around the benchmark: annotation pairing LLM
drafts with human correction, supervised and reinforcement training
pipelines, live evaluation with functional checkers and an LLM judge, and
joint metrics for task success and guide quality. Every successful run
yields a verified outcome and a candidate guide
(Figure~\ref{fig:teaser}(c)), reducing guide authoring to review.
Benchmarking three frontier API models and an open 9B model under this
protocol shows the task is far from solved: the best configuration
completes 37.4\% of the test tasks. It also shows that grounding is a real
design choice with no universal answer: Gemini is far stronger with
Set-of-Mark selection, GPT-5.5 and Claude show no meaningful preference,
and the trained 9B model improves only under Set-of-Mark grounding (13.2\%
versus 9.2\%).

Finally, we study how far training can push the small model. Supervised
finetuning on the gold demonstrations teaches the output format but not the
task: with Set-of-Mark grounding it reaches 6.9\%, below the 8.0\% of the
untuned base, because the tuned model learns to declare completion too early.
Plain GRPO~\cite{shao2024deepseekmath} then stalls: all-fail groups carry no
reward variance and yield no gradient, an issue also reported in large scale
RL systems~\cite{yu2026dapo}. Injecting cached expert trajectories from a
frontier model into the groups, in the spirit of off-policy
guidance~\cite{yan2026learning}, restores the signal: Set-of-Mark success
climbs from 6.9\% to 13.2\% (coordinate grounding does not benefit), guide
quality improves as well, and a pass@6 analysis shows new capability rather
than sharpened sampling. The remaining gap concentrates on long tasks: every
9B variant stays near zero beyond eight gold steps, where the API models
still complete 20 to 38\%.

We make three contributions.
\begin{itemize}
\item The MAG task and, to our knowledge, the first benchmark to unify
multistep web task execution with guide generation over screenshots, with two
grounding schemes and a human verified gold guide at every step.
\item A full harness from annotation to live evaluation and joint metrics,
used to benchmark frontier and open models, exposing model specific grounding preferences and a shared long horizon gap.
\item A GRPO recipe augmented with expert trajectories that nearly doubles
the Set-of-Mark success rate of a supervised 9B agent (6.9\% to 13.2\%) while
improving its guides, supported by paired task comparisons and a pass@k study.
\end{itemize}

\begin{figure*}[t!]
  \centering
  \includegraphics[width=0.9\textwidth]{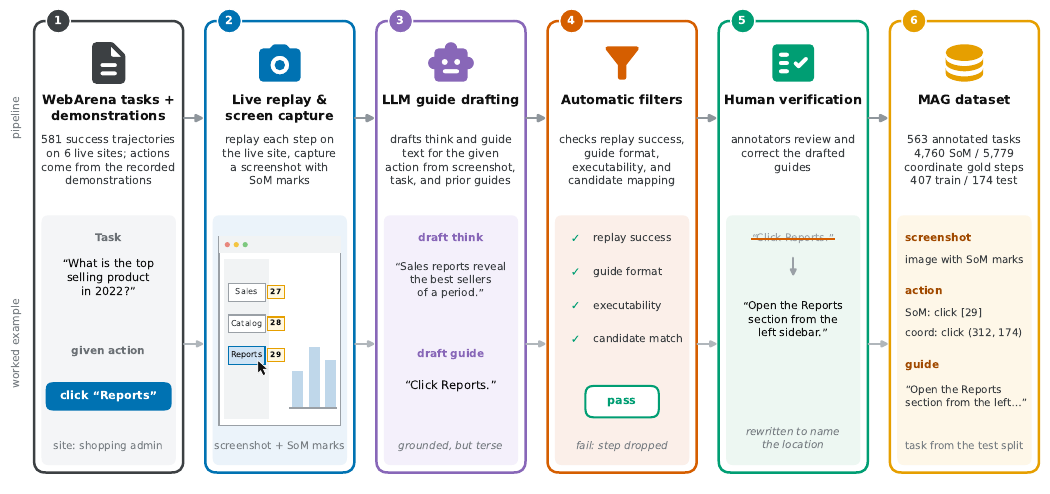}
  \caption{The MAG annotation pipeline (top) and one worked example step
  flowing through every stage (bottom). Failing steps are dropped; the
  407/174 split is defined over the 581 source tasks, of which 563 survive.}
  \label{fig:pipeline}
\end{figure*}

\section{Related Work}
\label{sec:related}

\noindent\textbf{Web agents from text to vision.}
Web agents have progressed from synthetic platforms~\cite{shi2017world} to
benchmarks that score functionally verified tasks on realistic
sites~\cite{yao2022webshop,deng2023mind2web,zhou2024webarena,drouin2024workarena,lu2024weblinx},
typically from a textual page abstraction on which strong agents layer
prompting or search~\cite{yao2022react,yang2025agentoccam,koh2024tree}. A
second line moves to what users see: Set-of-Mark prompting overlays
numbered marks on the screenshot~\cite{yang2023set}; SeeAct, WebVoyager,
CogAgent, and Pix2Act act from rendered pages or raw
pixels~\cite{zheng2024gpt,he2024webvoyager,hong2024cogagent,shaw2023pixels};
dedicated grounding models locate elements from
pixels~\cite{cheng2024seeclick,gou2025navigating}; and benchmarks now span
visual web tasks, desktops, and
phones~\cite{koh2024visualwebarena,xie2024osworld,rawles2025androidworld}.
In all of these the agent emits only actions and is judged by completion
alone.

\medskip
\noindent\textbf{Guide generation for web interfaces.}
Commercial DAPs attach human written guidance to production sites, and
keeping it aligned with evolving interfaces is a recognized maintenance
burden. Research on automating this is thin. \citet{gan2026guideweb}
formulate guide generation for a single given page: pick the element a user
should interact with and write the matching instruction. Nothing is executed,
so the guide is never verified against task success, and assistance stops at
one page.

\medskip
\noindent\textbf{Reinforcement learning for web agents.}
GRPO~\cite{shao2024deepseekmath,guo2025deepseek} made critic free
reinforcement learning practical for LLMs; prompted reflection improves
agents without weight updates~\cite{shinn2023reflexion}; WebRL and
WebAgent-R1 apply online RL to web
agents~\cite{qi2025webrl,wei2025webagent}. Two failure modes shape our
recipe. A group whose rollouts all fail carries zero advantage, which DAPO
counters with dynamic sampling~\cite{yu2026dapo}; and on-policy exploration
cannot discover what the policy cannot yet do, which LUFFY counters by mixing
off-policy expert traces into the groups~\cite{yan2026learning}. Binary
rewards from live websites make both problems severe for a 9B agent.

\medskip
\noindent\textbf{Our work.}
MAG differs from all three lines at the level of the task: the agent must act
and explain in the same step, and both outputs are scored. The coupling is
not cosmetic: a guide is trustworthy only if the step it describes advances
the task, and live execution is what verifies this. Against agent
benchmarks, MAG adds a second supervised output; against single page guide
generation, it covers whole tasks on live sites from the visual inputs
users see, and it offers, to our knowledge, the first controlled comparison
of Set-of-Mark and coordinate grounding on identical tasks. On the training
side, we carry off policy guidance to multimodal web RL and show it is the
enabling component: without expert traces, GRPO yields no lift here.

\section{The MAG Benchmark}
\label{sec:benchmark}

\subsection{Task Definition}
\label{sec:task}

An MAG episode is a triple $(q, s_0, \Phi)$: a natural language intent $q$,
an initial page state $s_0$ on one of six live websites, and a functional
checker $\Phi$ inherited from WebArena. At step $t$ the agent receives an
observation $o_t$ and the history of its own guides $g_{<t}$, and must
produce an action and a guide sentence jointly:
\begin{equation}
(a_t, g_t) = \pi_\theta\!\left(q,\, o_t,\, g_{<t}\right).
\label{eq:policy}
\end{equation}
The guide history is the only text state carried across steps: what the
agent tells the user is also what it remembers.

Both schemes receive the identical observation: a $1440\times900$ viewport
screenshot $x_t$ in which interactive elements are marked with numbered
boxes, and a candidate menu $C_t=\{(j,\, e_j,\, v_j)\}_{j=1}^{n_t}$ listing
each mark's element type $e_j$ and visible text $v_j$. The menu also lists
elements beyond the visible fold, which \texttt{scroll} brings into view,
so $o_t = (x_t, C_t)$ throughout and the input side is held fixed. The two
schemes differ only in how the action is grounded. An action is a triple
\begin{equation}
a_t = (\alpha_t,\, \rho_t,\, \omega_t),\qquad \alpha_t \in \mathcal{A},
\label{eq:action}
\end{equation}
with a verb $\alpha_t$ from the seven verb space
$\mathcal{A}=\{$\texttt{click}, \texttt{type}, \texttt{select},
\texttt{scroll}, \texttt{press\_enter}, \texttt{go\_back},
\texttt{finish}$\}$, a grounding argument $\rho_t$, and a payload $\omega_t$
(text to type, an option label, a scroll direction, or the final answer).
The schemes instantiate $\rho_t$ differently: $\rho_t \in \{1,\dots,n_t\}$
picks a mark under Set-of-Mark (SoM) grounding, while under coordinate
grounding $\rho_t$ is a pixel position on the page; $\rho_t$ is required
exactly for the element verbs \texttt{click}, \texttt{type}, and
\texttt{select}. Prompt, observation, budget, and scoring are all held
identical, so any performance difference between the schemes is
attributable to the grounding of the action itself.

The episode ends when the agent emits \texttt{finish} or after $H=25$ steps,
and success is judged functionally on the live site:
\begin{equation}
S = \Phi\!\left(s_{T+1},\, \omega_T\right) \in \{0, 1\},
\label{eq:success}
\end{equation}
where $T$ is the index of the last executed step, $s_{T+1}$ the final page
state, and $\omega_T$ the answer returned with \texttt{finish}; if the
budget expires without \texttt{finish}, $\omega_T$ is empty, so checkers
that require an answer score $0$ while state based checkers still evaluate
the final page. The defining constraint of MAG is the asymmetry
between the two outputs: the action may use marks or pixels, but the guide
$g_t$ must be a short instruction a user could follow on the visible page,
free of mark ids and coordinates. Every step is solved twice, in machine
terms and in human terms, and Section~\ref{sec:eval} scores both.

\subsection{Dataset Construction: LLM-Assisted Human Annotation}
\label{sec:dataset}

Section~\ref{sec:task} fixes an interface that no existing resource fills:
WebArena ships tasks and checkers, but no demonstrations in either grounding
form and no guide text at any step. We therefore build MAG on top of the
success demonstrations released by OpAgent~\cite{guo2026opagent}:
trajectories for 581 of WebArena's 812 tasks, recorded in coordinate form
and verified by the source authors, which we convert and enrich in the
pipeline of Figure~\ref{fig:pipeline}. MAG inherits exactly those tasks, so
the benchmark leans toward tasks a strong prior agent could complete, and
its success rates are not comparable to WebArena leaderboard numbers.

The first stage replays every step on the live sites and saves a
Set-of-Mark screenshot of the page as it looked at that step, mapping each
recorded coordinate onto the marked candidate it hits. One recorded
trajectory thus yields a coordinate view and a SoM view of the same
behavior, the duality behind the controlled grounding comparison of
Section~\ref{sec:exp}; the views share their tasks, though a step whose
target maps to no marked candidate survives only in the coordinate view
(Appendix~\ref{app:dataset}). The second stage adds the guides: a full web
task can be long and entangled, but writing the instruction for one step,
given the screenshot, the task, the action taken, and the guides so far, is
squarely within the competence of a frontier LLM, so
GPT-5.5~\cite{openai2026gpt55} drafts a think rationale and a guide
sentence for every step.

Because LLM drafting alone is not trustworthy, two review mechanisms
follow. A rule based filter drops steps with replay mismatches, malformed
guides, actions outside the seven verb space, or unmapped targets; the
survivors then pass a human stage in a review interface purpose built for
MAG (Appendix~\ref{app:review_tool}), where three annotators step through
every guide next to its screenshot and correct it on the spot. Their
consistent report: the drafts were largely correct, and few corrections
were needed. Every guide is human verified, and a usefulness check on 50
sampled test tasks confirms the references work in practice: 82\% let a
first time user complete the task (Appendix~\ref{app:review_tool}). The
result is 563 fully annotated tasks (396 training, 167
test) with a screenshot, both action forms, and a guide at every step,
4,760 SoM and 5,779 coordinate gold steps; the 407/174 split is fixed over
the 581 source tasks before capture. Statistics and licensing are in
Appendix~\ref{app:dataset}.

\subsection{Evaluation Suite}
\label{sec:eval}

MAG contributes more than a dataset: it comes with an evaluation suite
designed for the dual output of Equation~\ref{eq:policy}. Task success
alone says nothing about whether the produced guides could walk a user
through the task; text overlap alone rewards a fluent guide attached to a
failed trajectory. The suite therefore scores the task, the guides, and a
fused headline that credits guides only on solved tasks.

\paragraph{Task success.}
SR is judged on the live site by WebArena's functional checkers (exact
URL, page content, and program queries), applied to the final state and
answer as in Equation~\ref{eq:success}; 29 of the 174 test tasks need
semantic answer comparison and are scored by a fixed LLM judge that sees
only the answer and the reference.

\paragraph{Guide quality.}
For task $i$, the predicted guides $\hat g^{(i)}$ are joined in step order
and compared to the joined gold guides $g^{\star(i)}$ with four reference
metrics of equal weight, $\mathcal{M} = \{$BLEU-1, BLEU-2, ROUGE-1,
ROUGE-L$\}$~\cite{papineni2002bleu,lin2004rouge}:
\begin{equation}
G_i = \tfrac{1}{4}\!\sum_{m\in\mathcal{M}} m\!\left(\hat g^{(i)},
g^{\star(i)}\right)\!.
\label{eq:guideq}
\end{equation}

\paragraph{Gated Guide Score.}
The headline metric gates guide quality by success:
\begin{equation}
\mathrm{GGS}_i = S_i\left(\gamma + (1-\gamma)\, G_i\right),
\qquad \gamma = 0.4,
\label{eq:ggs}
\end{equation}
Success is scored on all 174 test
tasks; the guide bearing metrics average over the 171 test tasks with
reference guides, since capture failed entirely for three tasks and left
them without references. The gate
encodes the product requirement: a failed task scores zero regardless of
guide fluency, while on a solved task the guides modulate the score between
$\gamma$ and $1$. By
construction $\gamma\,\mathrm{SR} \le \mathrm{GGS} \le \mathrm{SR}$, and we
call the gap $\mathrm{SR} - \mathrm{GGS}$ the guide tax: the score a system
loses to imperfect guides. Because GGS is coupled to SR by design, we always
report SR and the ungated $G$ next to it.

\paragraph{Format gate.}
OFCR (Output Format Correctness Rate) is the fraction of steps that parse
under the output contract, form a valid action in the seven verb space with
the required arguments, and keep the guide free of leaked internals: a guide
that mentions a mark id or a pixel coordinate fails the gate, since it is
useless to a user who sees neither.

\paragraph{Step diagnostics and protocol.}
For training time analysis we additionally score teacher forced steps with
SAA (action correctness against gold) and GACS, a fused step score
$F \cdot A \cdot \sqrt{\mathrm{Faith}\cdot\mathrm{Suff}}$; its gold free
Faith and Suff terms double as guide diagnostics during training, while the
GRPO reward itself is binary task success (Section~\ref{sec:grpo}). Every
reported run follows one protocol: environment reset and fresh
authentication before each sweep, greedy decoding for locally served models
(API models use fixed provider settings), the same locked prompt, and the
25 step budget. Formal definitions and the evaluation
pseudocode are in Appendix~\ref{app:metrics}.

\begin{figure*}[t!]
  \centering
  \includegraphics[width=0.9\textwidth]{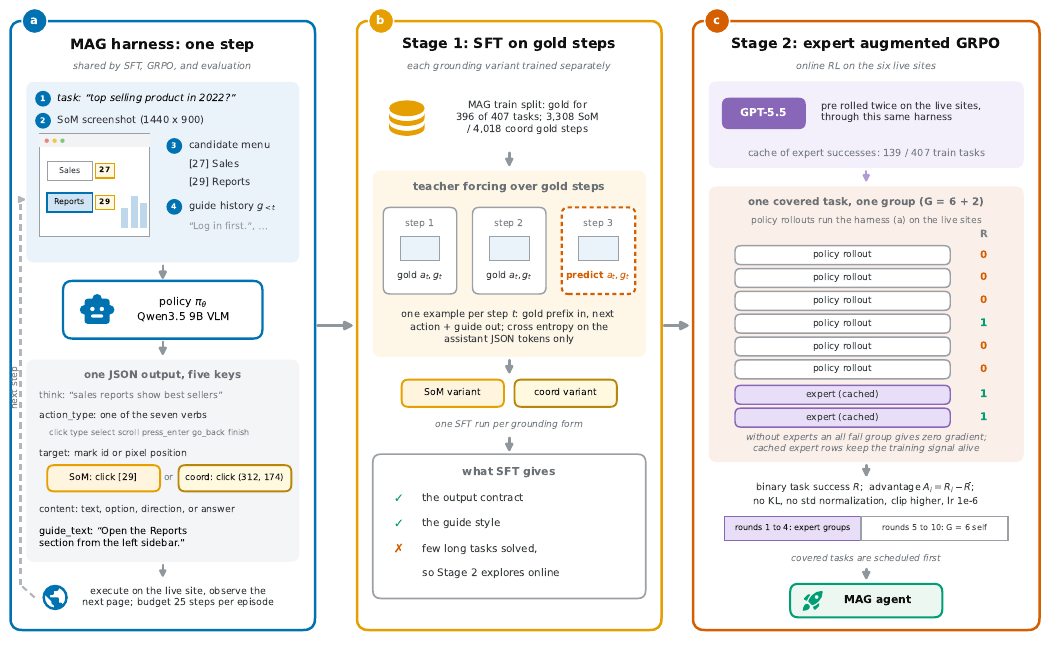}
  \caption{The MAG agent. (a) The shared harness, one step. (b) Stage 1
  SFT targets the output contract and the guide style. (c) Stage 2 expert
  augmented GRPO: GPT-5.5 is pre rolled twice and cached; covered tasks form
  groups of six policy plus two expert rollouts under a binary success
  reward.}
  \label{fig:agent}
\end{figure*}

\section{The MAG Agent}
\label{sec:agent}

\subsection{The MAG Harness}
\label{sec:arch}

Everything in this paper, the three API baselines, the SFT corpus, every
GRPO rollout, and every reported number, runs through one harness
(Figure~\ref{fig:agent}a). Its job is to turn a live website and a vision
language model into a loop stable enough to train against and fair enough
to compare across models; much of the difficulty of MAG lives here, and we
release it in full.

On the input side, each step renders the page in a $1440\times900$
viewport, marks the interactive elements, and builds the candidate menu of
Section~\ref{sec:task}, up to 130 lines of \texttt{[id] TYPE visible text}
with elements beyond the fold included. The locked system prompt fixes the
output contract: the five JSON keys, the seven verbs with per verb
parameter rules, general operating rules that reference no specific site,
page, or answer, and the guide rules that ban mark ids and coordinates from
user facing text. The user message carries the task, the guide history, the menu, and the
screenshot, under budgets of 16k input and 1{,}024 output tokens. The full prompt and a
complete worked step are reproduced in Appendix~\ref{app:prompts}.

On the output side, one tolerant parser is shared by data construction,
training, and evaluation: it extracts the first well formed JSON object
from the raw completion, surviving code fences and reasoning preambles. The
executor maps the seven verbs onto browser primitives on the live page; a
step that yields no valid action executes nothing and consumes budget.
Because parser and executor are identical everywhere, OFCR measures
exactly the gate that training and evaluation apply.

The operational layer is where live web RL usually breaks, and each rule
here exists because its absence corrupted an experiment: six replicated
environment sets run the rollouts of a group in parallel; every round and
every sweep begins by resetting the site containers to their snapshots and
re registering, then verifying, every account, since stale logins silently
depress success rates; judge calls queue through a gateway when the
training host has no API egress; and every constant is frozen in one
configuration module that all components import, so no two stages can
drift apart.

\subsection{Stage 1: SFT for the Contract and the Guide Style}
\label{sec:sft}

Stage 1 finetunes the base VLM~\cite{qwen2026omni} separately for each
grounding form on its gold steps, 3{,}308 SoM and 4{,}018 coordinate
examples over the 396 annotated training tasks, each example rendered
exactly as at inference and supervised with cross entropy on the assistant
JSON tokens only (full finetuning; hyperparameters in
Appendix~\ref{app:training}).

The purpose of this stage is deliberately not task competence; it is
twofold. First, the contract. The base model produces a parseable,
executable step in only 19\% of attempts under SoM grounding and 14\% under
coordinates, and nothing downstream survives that. After SFT the rate is
about 97\% for both variants, and every later stage presumes it. Second, the guide register. The gold guides
carry the imperative, user facing style that the MAG task demands, and this
stage is where that style is learned; the Stage 2 reward never scores
guides, yet guide quality persists and improves through RL precisely
because SFT anchored it. What SFT does not deliver is success: rates barely
move, and the SoM variant even trails its base by learning to declare
completion too early (Section~\ref{sec:results}). Competence is the next
stage's job.

\subsection{Stage 2: Expert Augmented GRPO}
\label{sec:grpo}

With a binary reward on a live site, GRPO~\cite{shao2024deepseekmath}
learns only from groups whose rollouts disagree: every trajectory in an
all fail group has zero advantage. At SFT level competence this is the
common case: across ten plain GRPO attempts spanning reward shaping,
curricula, and penalty terms, no run produced a sustained gain
(Appendix~\ref{app:failedruns}); dynamic sampling in the style of
DAPO~\cite{yu2026dapo} does not help, because it assumes solvable prompts
exist to be resampled, while here the policy never reaches a first success
on most tasks. The bottleneck is capability, not variance reduction.

We therefore import the missing successes from outside the policy, in the
spirit of off policy guidance~\cite{yan2026learning}. Before training we
roll GPT-5.5~\cite{openai2026gpt55} through the same harness on the 407
training tasks twice, two independent passes under identical budgets, and
cache only the trajectories the evaluator verifies as successful: 112
and 122 tasks, 139 in union (95 in both). A cached trajectory stores the
full step records, so it enters a group exactly like a policy rollout.

For a covered task $q$, a group joins six on policy rollouts, sampled on
the live environments at temperature 1.0 under the 25 step budget, with the
cached expert trajectories $\mathcal{E}_q$ ($|\mathcal{E}_q| \le 2$):
\begin{equation}
\mathcal{G}_q = \{\tau_1,\dots,\tau_6 \sim \pi_{\theta_{\mathrm{old}}}\}
\cup \mathcal{E}_q,
\quad
R_i = S(\tau_i),
\label{eq:group}
\end{equation}
with $S \in \{0,1\}$ the functional success of Equation~\ref{eq:success}.
Advantages are centered but not rescaled:
\begin{equation}
A_i = R_i - \frac{1}{|\mathcal{G}_q|}\sum_{j} R_j;
\label{eq:adv}
\end{equation}
groups with zero reward variance are dropped, and no standard deviation
division is applied, so the advantage of a lone success in a failing group
keeps its full magnitude. The policy ascends the token level clipped
objective~\cite{schulman2017proximal}
\begin{equation}
J(\theta) = \frac{1}{\sum_i |\tau_i|} \sum_{i,t}
\min\!\big(\rho_{i,t} A_i,\; \tilde\rho_{i,t} A_i\big),
\label{eq:grpoobj}
\end{equation}
where $\rho_{i,t}$ is the importance ratio between $\pi_\theta$ and
$\pi_{\theta_{\mathrm{old}}}$ on token $t$ of $\tau_i$ and
$\tilde\rho_{i,t} = \mathrm{clip}(\rho_{i,t},\, 1-\varepsilon_l,\,
1+\varepsilon_h)$ with an asymmetric clip $\varepsilon_l = 0.2$,
$\varepsilon_h = 0.28$~\cite{yu2026dapo}, which throttles positive
advantages less; no KL term is applied.

Two details matter. The teacher's token probabilities are unavailable,
so $\rho$ on expert tokens is computed against the policy's own round start
log probabilities, making expert rows ordinary off policy data with
$A_i > 0$. And round start log probabilities are recomputed once per round
with the training stack, not the inference engine, whose numerics differ;
the frozen $\pi_{\theta_{\mathrm{old}}}$ keeps the clipped objective
meaningful across a round's updates.

Training runs ten rounds, expert covered tasks first, so early rounds see
expert groups and later rounds continue with plain six rollout groups; each
round resets and re authenticates the environments, collects rollouts,
merges judge verdicts, and performs one inner epoch of updates
(Algorithm~\ref{alg:grpo}, Appendix~\ref{app:training}). The reward carries
no guide term, and yet guide quality improves alongside success
(Section~\ref{sec:coupling}): the style anchored in Stage 1 rides along
with competence.

\begin{table*}[t]
\centering
\small
\resizebox{0.86\textwidth}{!}{%
\begin{tabular}{llrrrrrrrrr}
\toprule
Model & Grounding & SR & $N_{\mathrm{succ}}$ & OFCR & GGS & TAX & BLEU-1 & BLEU-2 & ROUGE-1 & ROUGE-L \\
\midrule
\multicolumn{11}{l}{\textit{API models, no finetuning}} \\
GPT-5.5           & SoM   & .356 & 62 & \textbf{\underline{.999}} & .206 & .150 & .236 & .168 & .340 & .269 \\
GPT-5.5           & coord & \textbf{\underline{.374}} & \textbf{\underline{65}} & .995 & \textbf{\underline{.225}} & .149 & \textbf{\underline{.276}} & \textbf{\underline{.200}} & \textbf{\underline{.380}} & \textbf{\underline{.294}} \\
Gemini 3.5 Flash  & SoM   & .345 & 60 & .949 & .200 & .145 & .236 & .157 & .326 & .251 \\
Gemini 3.5 Flash  & coord & .207 & 36 & .964 & .103 & .104 & .148 & .088 & .214 & .178 \\
Claude Sonnet 4.6 & SoM   & .276 & 48 & .883 & .154 & .122 & .194 & .128 & .265 & .206 \\
Claude Sonnet 4.6 & coord & .270 & 47 & .904 & .145 & .125 & .181 & .117 & .256 & .202 \\
\midrule
\multicolumn{11}{l}{\textit{Qwen3.5 9B, ours}} \\
base              & SoM   & .080 & 14 & .186 & .049 & .032 & .199 & .138 & .301 & .245 \\
base              & coord & .017 & 3  & .137 & .003 & .014 & .156 & .097 & .256 & .208 \\
SFT$^\dagger$     & SoM   & .069 & 12 & .976 & .036 & .033 & .219 & .153 & .306 & .250 \\
SFT               & coord & .075 & 13 & .974 & .044 & .031 & .237 & .182 & .327 & .271 \\
GRPO round 5      & SoM   & .109$^{\uparrow}$ & 19 & .996 & .064$^{\uparrow}$ & .045 & .248 & .185 & .358 & \textbf{\underline{.284}} \\
GRPO round 10     & SoM   & \textbf{\underline{.132}}$^{\uparrow}$ & \textbf{\underline{23}} & .978 & \textbf{\underline{.076}}$^{\uparrow}$ & .056 & .256 & .188 & \textbf{\underline{.358}} & .280 \\
GRPO round 5      & coord & .092 & 16 & .998 & .054 & .038 & \textbf{\underline{.260}} & \textbf{\underline{.191}} & .344 & .281 \\
GRPO round 10     & coord & .080 & 14 & \textbf{\underline{.998}} & .052 & .028 & .240 & .178 & .330 & .271 \\
\bottomrule
\end{tabular}}
\caption{Main results under the protocol of Section~\ref{sec:eval}. SR
counts successes over all 174 test tasks; the guide metrics (GGS, TAX,
BLEU, ROUGE) average over the 171 test tasks with reference guides
(Section~\ref{sec:eval}). OFCR is recomputed from raw outputs under the
Appendix~\ref{app:metrics} definition. Bold underlined values mark the best
per column within each panel; $\uparrow$ marks the SoM GRPO gains over
their SFT anchor. $^\dagger$The same weights in an independent sweep
measure .063; we report the value from inside the GRPO run, which anchors
the round rows.}
\label{tab:main}
\end{table*}

\section{Experiments}
\label{sec:exp}

\subsection{Setup and Baselines}
\label{sec:setup}

We evaluate three frontier API models, GPT-5.5~\cite{openai2026gpt55},
Gemini 3.5 Flash~\cite{google2026gemini}, and Claude Sonnet
4.6~\cite{anthropic2026sonnet}, and the open Qwen3.5 9B
VLM~\cite{qwen2026omni}\footnote{\url{https://huggingface.co/Qwen/Qwen3.5-9B};
our SFT and GRPO checkpoints are released with the harness.} as base, after
SFT, and after GRPO rounds 5 and 10, each under both grounding schemes.
Every run follows the protocol of Section~\ref{sec:eval} with the same
prompt, parser, budgets, and judge.

We deliberately compare against no earlier WebArena agent. MAG is a new
task with a second scored output; its tasks are reannotated and inherited
from one agent's solvable subset (Section~\ref{sec:dataset}); and its
metrics require guides that existing agents do not produce. The evaluation
therefore answers how well current models do MAG, not where they rank on
WebArena.

\subsection{Main Results}
\label{sec:results}

Table~\ref{tab:main} reports the full suite; four findings organize what follows.

\paragraph{MAG is far from solved.}
The best configuration, GPT-5.5 with coordinates, completes 37.4\% of the
test tasks; the best trained 9B agent reaches 13.2\%. Every model also pays a guide tax: even the best GGS (.225) sits far below its own SR.

\paragraph{The contract is learnable; competence is not the same thing.}
API models satisfy the output contract out of
the box (OFCR .88 to .999); the 9B base does not (.19 SoM, .14 coord), and
SFT repairs exactly this, .97 or higher on every tuned row
(Section~\ref{sec:sft}). No evaluated step in any run leaked a mark id or coordinate into a guide:
the constraint is easy to satisfy; being right is not.

\paragraph{Grounding preference is model specific.}
Gemini is far stronger with SoM than with coordinates (+13.8 points; 34
test tasks solved only with SoM against 10 only with coordinates), GPT-5.5
and Claude show no meaningful preference (14 against 11, and 13 against
12), and the trained 9B agent makes progress only under SoM: 6.9 to
13.2 through GRPO, while its coordinate variant peaks at 9.2 in round 5 and
falls back to 8.0 by round 10: grounding has to be measured per model,
exactly the comparison MAG enables.

\paragraph{Expert augmented GRPO is the only recipe that moves success.}
Under SoM it lifts the same
weights from 6.9 to 10.9 to 13.2 (+6.3 points over SFT; 17 tasks gained
against 6 lost), nearly doubling success and more than doubling GGS (.036
to .076), with the
reference guide metrics rising alongside although the reward never scores
guides; under coordinates it does not, consistent with the grounding
finding.

\section{Analysis}
\label{sec:analysis}

\paragraph{How solid are the gains.}\label{sec:significance}
With 174 test tasks, single digit gaps deserve scrutiny. The headline SoM
gain is +6.3 points, 17 tasks solved only by GRPO against 6 only by SFT
(task bootstrap 95\% CI $[+1.1, +11.5]$); the intermediate steps are
monotone though individually within noise. Coordinate gains are small (CI
$[-3.4, +4.6]$ at round 10), and the SoM minus coordinate gain difference,
+5.7 points, still crosses zero ($[-0.6, +12.1]$), so we describe SoM as
the mode where training makes progress rather than claim a proven
contrast. Gemini's SoM advantage is the largest modality effect (+13.8,
CI $[+6.9, +21.3]$); GPT-5.5 with coordinates leads the best 9B agent by
+24.1 (CI $[+17.2, +31.6]$). Full table: Appendix~\ref{app:fullresults}.

\paragraph{Sharpening or learning? A pass@k view.}\label{sec:passk}

Sampling six trajectories per task at temperature 1.0 separates two
readings of the GRPO gain. If RL only sharpened the SFT distribution,
pass@6 would stay flat; instead it rises from 14.4\% (SFT) to 21.3\% (round
10), +6.9 points (21 tasks solved only by GRPO
against 9 only by SFT; CI $[+1.2, +13.2]$), alongside pass@1 (4.9 to
12.1). Sampling hurts the SFT policy (greedy 6.9 versus sampled pass@1 4.9),
while the GRPO policy stays sharp (12.1 versus 13.2 greedy). GRPO adds capability, not just concentration (Appendix~\ref{app:pass6}).

\paragraph{What GRPO learns.}\label{sec:whatlearned}

Solved sets: round 10 solves 17 tasks SFT
could not and loses 6, so the lift is new competence rather than variance
reduction; the union of its SoM and coordinate successes covers 16.7\% of
tasks (overlap only 8), an easy routing headroom. Actions: scroll drops from 34.9\% to 13.5\% of steps, click rises to
57.9\%, and go\_back reappears: timid wandering turns into decisive
interaction. Horizon: gains concentrate on
short and medium tasks (+9.8 and +9.1 points), while beyond eight gold steps
every 9B variant stays under 3\%, where the API models sustain 20 to
38\% (Appendix~\ref{app:fullresults}). The remaining teacher gap is a long
horizon gap, not a per step accuracy gap.

\paragraph{Ablation: expert injection.}\label{sec:ablation}

Ten earlier plain GRPO configurations without expert rows produced no
sustained gain (first to last round deltas $-14.6$ to $+1.0$ points),
while the number of groups carrying reward variance tracks expert coverage
almost exactly; Appendix~\ref{app:failedruns} documents the runs and the
observational caveats.

\paragraph{Guide and success coupling.}\label{sec:coupling}

The two outputs of the task are not independent. On round 10, guide
quality $G$ on solved episodes is .369 against .256 on failed ones; after
SFT the gap is .351 against .225. Both gaps are stable under task level
resampling (95\% CIs $[+.03, +.19]$ and $[+.02, +.22]$, over the 171
referenced tasks), and the coupling strengthens through RL although the
reward never sees a guide.
The agent that can do the task describes it better, the premise of
unifying the two outputs (Appendix~\ref{app:metrics}).

\section{Conclusion}
\label{sec:conclusion}

MAG turns the manual labor behind in app web guidance into a measurable
task: complete a live multistep web task from screenshots and write, at
every step, the guide a user would need. We contribute the benchmark, the end to end harness, an evaluation
suite that credits guides only on solved tasks, and a training
recipe in which cached expert trajectories restore the signal plain GRPO
lacks. The best frontier
configuration completes 37.4\% and small models fail on long horizons; we
release everything to make progress on MAG measurable.

\section*{Limitations}

MAG is a new and deliberately hard task: live multistep websites,
screenshot only observation, a 25 step budget, and two jointly scored
outputs. Absolute success rates are accordingly low, 37.4\% for the
strongest frontier configuration and 13.2\% for our best 9B agent, and
should be read as a measure of the task's difficulty and headroom rather
than of any single method; the benchmark exists to make progress on this
gap measurable. Beyond that, the usual caveats of a first release apply.
The test set holds 174 tasks, so single digit differences carry roughly
$\pm 2$ point uncertainty, and the GRPO result rests on one seed per
grounding scheme and one teacher model. The guide metric is single
reference (Appendix~\ref{app:metrics}), all tasks come from six
WebArena sites, and train and test tasks largely share WebArena intent
templates (Appendix~\ref{app:fullresults}), so transfer beyond seen
templates and sites remains to be shown.

\section*{Ethics Statement}

All sites are self hosted WebArena sandboxes populated with synthetic
content; no live third party service is acted upon and no personal data is
collected or processed. Guide annotation and verification were carried out
by three members of the research team. API models were accessed under
their providers' terms of service; the released corpus and harness carry
the Apache 2.0 license, and released screenshots contain sandbox content
derived from Wikipedia and OpenStreetMap, redistributed under their
respective terms (Appendix~\ref{app:dataset}). The intended use of MAG is
assistive: reducing the authoring cost of in app guidance. We see limited
misuse potential beyond generic web automation concerns, which the
sandboxed environment does not enable.



\bibliography{custom}

\appendix
\raggedbottom  
\setcounter{table}{0}
\renewcommand{\thetable}{A\arabic{table}}
\setcounter{figure}{0}
\renewcommand{\thefigure}{A\arabic{figure}}

\section{Dataset Details}
\label{app:dataset}

\begin{table}[H]
\centering
\small
\resizebox{\columnwidth}{!}{%
\begin{tabular}{lrrrrr}
\toprule
Site & Train & Test & Tasks & \multicolumn{2}{c}{Gold steps} \\
\cmidrule(lr){5-6}
 & & & & SoM & Coord \\
\midrule
GitLab          & 99  & 41  & 140 & 1,286 & 1,664 \\
Map             & 53  & 23  & 76  & 599   & 606   \\
Reddit          & 69  & 29  & 98  & 502   & 583   \\
Shopping        & 78  & 33  & 111 & 659   & 858   \\
Shopping admin  & 88  & 38  & 126 & 1,644 & 1,998 \\
Wikipedia       & 9   & 3   & 12  & 70    & 70    \\
\midrule
Total           & 396 & 167 & 563 & 4,760 & 5,779 \\
\bottomrule
\end{tabular}}
\caption{Per site composition of the annotated MAG corpus. Train and test
counts refer to the 563 tasks that survive capture and filtering; the
407/174 split itself is defined over all 581 source tasks.}
\label{tab:dataset_sites}
\end{table}

\begin{table}[H]
\centering
\small
\begin{tabular}{lrrrr}
\toprule
Verb & \multicolumn{2}{c}{SoM} & \multicolumn{2}{c}{Coord} \\
\cmidrule(lr){2-3}\cmidrule(lr){4-5}
 & \# & \% & \# & \% \\
\midrule
\texttt{click}        & 2,572 & 54.0 & 3,392 & 58.7 \\
\texttt{type}         & 1,057 & 22.2 & 1,242 & 21.5 \\
\texttt{finish}       & 563   & 11.8 & 563   & 9.7  \\
\texttt{scroll}       & 415   & 8.7  & 415   & 7.2  \\
\texttt{select}       & 120   & 2.5  & 134   & 2.3  \\
\texttt{go\_back}     & 33    & 0.7  & 33    & 0.6  \\
\texttt{press\_enter} & 0     & 0.0  & 0     & 0.0  \\
\midrule
Total                 & 4,760 &      & 5,779 &      \\
\bottomrule
\end{tabular}
\caption{Action verb distribution of the gold steps in the two grounding
views.}
\label{tab:dataset_actions}
\end{table}

MAG is built from the success demonstrations released by
OpAgent~\cite{guo2026opagent} under the Apache 2.0 license, covering 581 of
WebArena's 812 tasks, replayed and
reannotated as described in Section~\ref{sec:dataset}. All pages are
rendered in a $1440\times900$ viewport with Set-of-Mark marking; the
candidate menu lists each mark as \texttt{[id] TYPE visible text}, with
visible text clipped to 80 characters, and includes elements beyond the
visible fold, which the \texttt{scroll} verb brings into view.
Table~\ref{tab:dataset_sites} breaks the corpus down by site,
Table~\ref{tab:dataset_actions} by action verb, and
Table~\ref{tab:guide_examples} shows gold guides spanning sites and verbs.

The two views share the same 563 tasks but differ in step counts for two
reasons. The SoM view drops a step when its recorded coordinate maps to no
marked candidate: 1,019 of the 4,768 element steps (21.4\%) survive only in
the coordinate view. In both views every trajectory ends with exactly one
\texttt{finish} step that carries the task answer and a templated guide
announcing it, synthesized when the recording lacked an explicit final
step, which is why the \texttt{finish} count equals the task count.
Trajectories average 10.3 steps in the coordinate view and 8.5 in the SoM
view (median 6 and maximum 71 in both), and gold guides average 14 words
(median 13). Eleven test tasks have gold demonstrations longer than the 25
step budget; the budget is identical for every system, so comparisons are
unaffected, though success along the demonstrated path is capped near
94\%. The geometric coordinate to candidate mapping that produces the SoM view
can pick the wrong mark when a flyout menu overlays the page; an anchor
based audit flags 3.9\% of gold SoM click targets as suspect (the guide
text strongly matches a different candidate), concentrated in such
menus. The affected steps are unaffected in the coordinate view. The verb \texttt{press\_enter} is available to the agent at inference
time but absent from the gold demonstrations, which submit forms through
visible controls instead. All sites are self hosted WebArena sandboxes
populated with synthetic content, so the corpus contains no real user data.
The corpus and the harness are released under Apache 2.0, matching the
license of the source demonstrations; released screenshots contain sandbox
content derived from Wikipedia (CC BY SA) and OpenStreetMap (ODbL),
redistributed under those terms.

\begin{table}[H]
\centering
\small
\resizebox{\columnwidth}{!}{%
\begin{tabular}{llp{4.1cm}}
\toprule
Site & Verb & Gold guide \\
\midrule
Shopping admin & \texttt{click} & Click the Reports menu item in the left
sidebar to open sales reporting options. \\
GitLab & \texttt{type} & Type `kilian' into the Author search field to find
Kilian's commits. \\
Map & \texttt{select} & Select Foot (OSRM) as the travel mode to calculate
the walking route time. \\
Reddit & \texttt{scroll} & Scroll down to review more comments and compare
each comment's upvote and downvote counts. \\
\bottomrule
\end{tabular}}
\caption{Gold guide examples across sites and verbs.}
\label{tab:guide_examples}
\end{table}

\section{Guide Verification Interface}
\label{app:review_tool}

\begin{figure*}[p]
  \centering
  \includegraphics[width=0.94\textwidth]{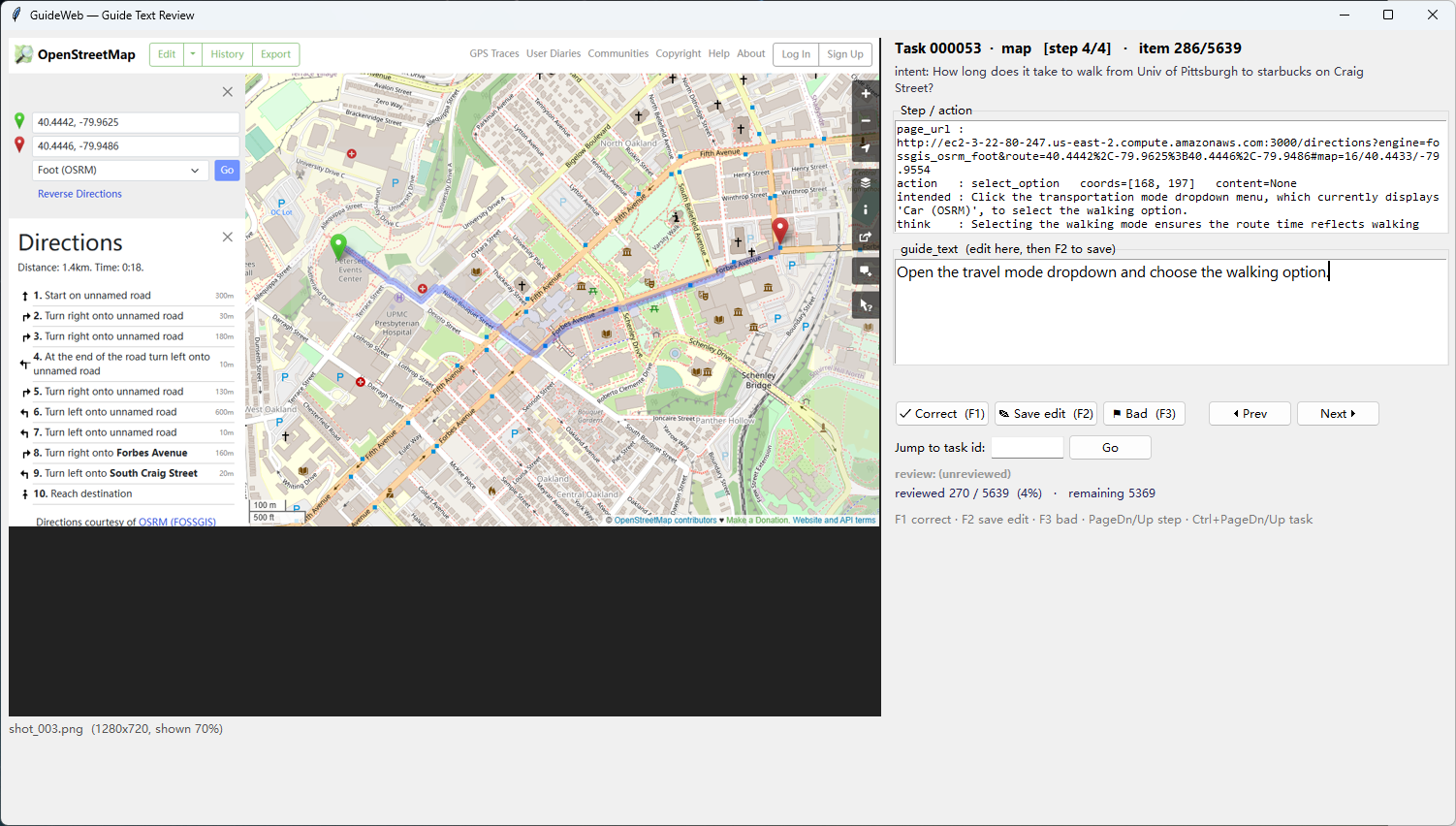}\\[6pt]
  \includegraphics[width=0.94\textwidth]{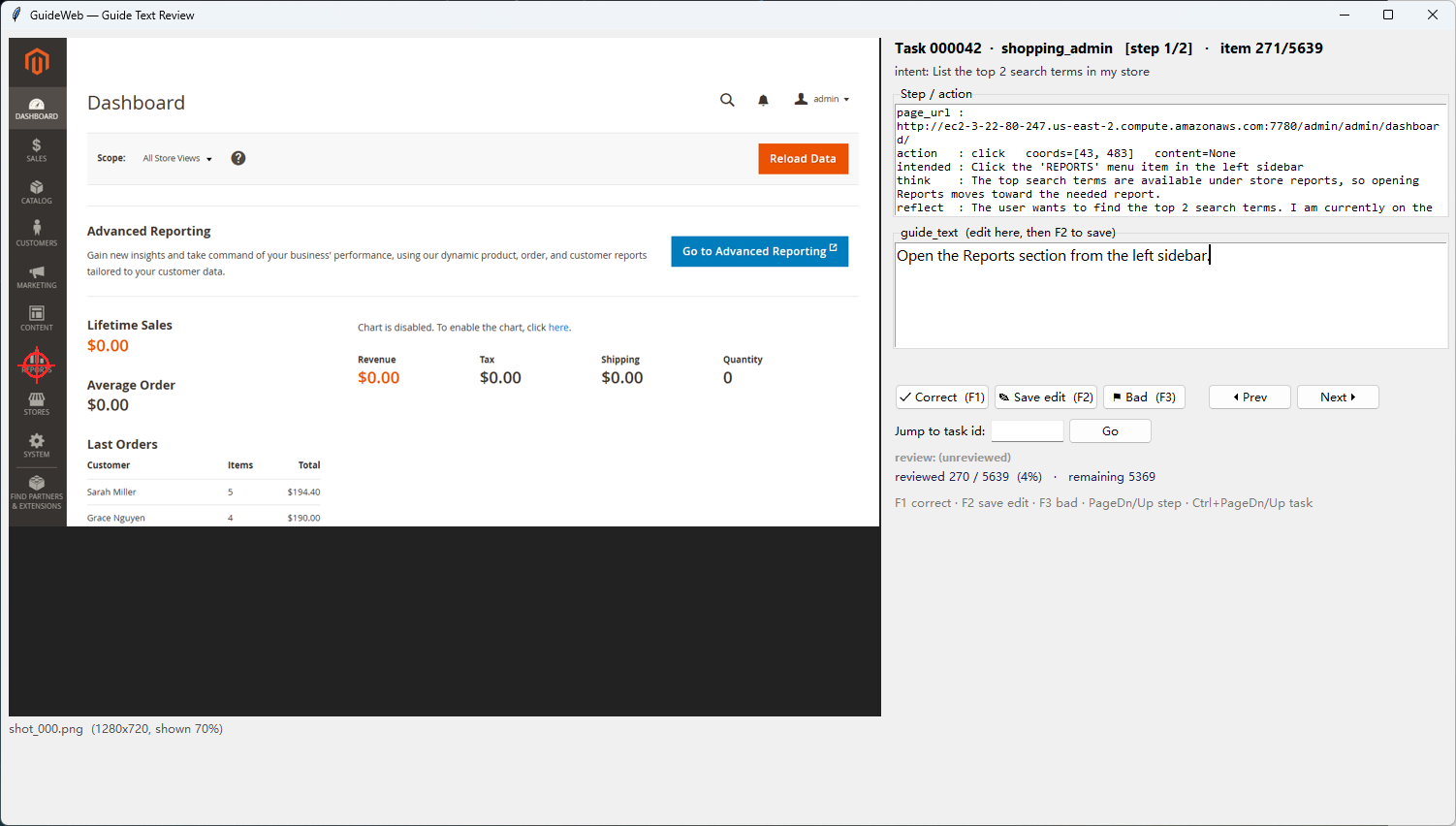}
  \caption{The guide verification interface on two annotation steps:
  selecting the walking mode for a route query on the map site (top) and
  opening the Reports section for a store analytics task on the shopping
  admin site (bottom). The main pane shows the captured screenshot that the
  LLM saw when drafting the step, and the right pane holds the task intent,
  the full step record, and the drafted guide sentence in an editable box.}
  \label{fig:review_ui}
\end{figure*}

Every drafted guide passes through a manual verification stage carried out
in the desktop tool shown in Figure~\ref{fig:review_ui}. The tool walks the
annotator through the dataset step by step. For each step it displays the
captured screenshot that served as the drafting input and lists the task
intent together with the step record: page URL, executed action, and the
drafted think. Guides were drafted and reviewed on the initial replay
captures ($1280\times720$, as visible in Figure~\ref{fig:review_ui}); the
released corpus recaptures every step at $1440\times900$ and realigns the
recorded actions. The guide sentence sits in
an editable text box, and three keyboard commands cover the entire workflow:
accept the guide as written, save an edited version, or flag the step as
bad. A progress counter tracks coverage and an index field jumps to any
task.

The review concentrates on the guides: the annotator checks that each
sentence names the correct target on the current page and reads as a clear,
self contained instruction for that step. Since all trajectories originate
from verified success demonstrations, action correctness is only spot
checked in passing. Most drafts passed review unchanged; typical edits added
the location of the target element or simplified wording.

\paragraph{Usefulness check.} To confirm that the verified reference guides
are useful to a person and not merely n-gram targets, one author rated the
step by step gold guide of 50 randomly sampled test tasks (stratified across
the six sites, fixed seed) as \emph{useful}, \emph{ambiguous}, or
\emph{useless}, where \emph{useful} means a first time user could complete
the task by following the guide alone. The outcome is 41 useful (82\%), 2
ambiguous (4\%), and 7 useless (14\%), and the useful rate has a Wilson
95\% interval of $[69.2\%,\,90.2\%]$. It is high on the shopping admin and
map sites (100\%), shopping (90\%), and GitLab (83\%), and lower on Reddit.
The incomplete cases are almost entirely a Set of Mark step mapping effect
rather than wording errors: when a terminal control such as a Post or Submit
button does not fall on a marked candidate, that step is dropped from the
Set of Mark sequence (the 21.4\% coordinate only element steps of
Appendix~\ref{app:dataset}), so the guide reads correctly only up to its
last mapped step. The per step guide text itself is human verified, and
because this mapping applies identically to the gold reference for every
model under a grounding scheme while success is scored by live functional
checkers independent of the guide, model comparisons are unaffected. This
check validates the LLM assisted and human verified annotation layer on the
released guide text.

\section{Prompts}
\label{app:prompts}

This appendix reproduces the locked prompt verbatim. One system prompt
(Figure~\ref{fig:sysprompt}) and one user template
(Figure~\ref{fig:usermsg}) serve the API baselines, SFT data construction,
GRPO rollouts, and evaluation; only the grounding argument differs between
the two variants.

\begin{figure*}[p]
\centering
{\setlength{\fboxsep}{0pt}\setlength{\fboxrule}{0.9pt}%
\fcolorbox[gray]{0.45}{1}{\begin{minipage}{0.96\textwidth}
{\setlength{\fboxsep}{5pt}%
\colorbox[gray]{0.32}{\makebox[\dimexpr\linewidth-10pt][l]{%
\textcolor{white}{\bfseries\small System Prompt (SoM variant)}}}}\par\nointerlineskip
{\setlength{\fboxsep}{8pt}%
\colorbox[gray]{0.97}{\begin{minipage}{\dimexpr\linewidth-16pt}
\scriptsize\ttfamily\raggedright
\begin{multicols}{2}
You are a vision web agent. You operate a real website one step at a time
to accomplish a user's task, and at every step you also produce one short
piece of in-app guidance for a human who is doing the same step.
\par\medskip
At each step you are given:
\begin{enumerate}\itemsep1pt\parskip0pt\topsep1pt
\item The user's task (the goal to accomplish).
\item A screenshot of the current page. Interactive elements are marked
with numbered boxes (Set-of-Marks); each number is the id of one candidate
element.
\item A numbered list of candidate elements, one per line, formatted as:
{}[id] TYPE visible text (for example: [17] BUTTON Add to Cart). Each id
matches a numbered box in the screenshot.
\item The history of guidance you have already produced this episode (what
has been done so far).
\end{enumerate}
Decide the single best NEXT action, then write its guide\_text.
\par\medskip
{\bfseries OUTPUT CONTRACT}\par
Return exactly one valid JSON object and nothing else: no markdown, no code
fences, no comments, no text before or after, and no second JSON object.
The object must have exactly these five keys:
\begin{itemize}\itemsep1pt\parskip0pt\topsep1pt
\item "think": a brief, concrete reasoning string for why this action
advances the task. This is your private reasoning and is not shown to the
human.
\item "action\_type": exactly one of: "click", "type", "select", "scroll",
"press\_enter", "go\_back", "finish". No other value is allowed.
\item "selected\_candidate\_id": the id (as a string) of the target element
from the candidate list, or null.
\item "content": the action's payload, or null (see the per-action rules
below).
\item "guide\_text": one short instruction (one sentence) telling a HUMAN
what to do this step.
\end{itemize}
{\bfseries ACTION SPACE} - use only these seven action types, and follow
each parameter rule exactly:
\begin{itemize}\itemsep1pt\parskip0pt\topsep1pt
\item "click": selected\_candidate\_id = the target id; content = null.
\item "type": selected\_candidate\_id = the input field id; content = the
exact text to type.
\item "select": selected\_candidate\_id = the dropdown id; content = the
exact visible option label to choose.
\item "scroll": selected\_candidate\_id = null; content = "up" or "down".
\item "press\_enter": selected\_candidate\_id = null; content = null. Use
this to submit the field you just typed into when no submit button is
needed.
\item "go\_back": selected\_candidate\_id = null; content = null.
\item "finish": selected\_candidate\_id = null; content = the final answer
if the task asks for one, otherwise null.
\end{itemize}
{\bfseries GENERAL OPERATING RULES} (these describe how any web UI behaves;
they never assume a specific site, page, or answer):
\begin{itemize}\itemsep1pt\parskip0pt\topsep1pt
\item For "click", "type", and "select", selected\_candidate\_id MUST be
one of the ids present in the candidate list for this step. Never invent an
id, and never target an element that is not listed.
\item Choose the element that most directly advances the task; prefer the
actionable control (a button, link, input, or dropdown) over a surrounding
container.
\item For "select", content must be an option label that actually exists in
that dropdown; do not invent option names. If the dropdown's options are
not visible yet, click to open it first.
\item After you type a query into a search or filter field, submit it
(press\_enter, or click the visible search/apply button) and let the
results update BEFORE you read any result or finish. Do not read a value
from, or finish on, a field you have only typed into but not yet submitted.
\item After setting filters, dates, periods, or options on a results or
report page, apply or run them so the results reflect your settings before
you read or finish.
\item Finish as soon as the information the task asks for is clearly
visible on the current page; do not take extra navigation steps once the
answer is on screen.
\item When the answer comes from on-screen text (a table cell, label,
heading, or field), copy it from the screenshot exactly as displayed:
preserve spacing, capitalization, punctuation, hyphens, numbers, units, and
any trailing qualifiers. Do not paraphrase, normalize, reorder, translate,
or complete it from prior knowledge.
\item For a task that only asks you to display, view, sort, or filter a
list, finish once the requested list or view is visible; do not open
individual items one by one unless the task needs a detail that is not
visible in the list.
\item For a task that asks for a single value (a name, count, price, date,
etc.), make the finish content concise - just the requested value, not a
full sentence.
\item Do not repeat an action that produced no useful change. In
particular, do not scroll repeatedly: if a recent step already scrolled and
nothing new and relevant appeared, act on a visible control instead, or
finish if the answer is now visible.
\item If a page requires signing in and credentials are available to you,
fill the fields once and submit; do not retype the same login repeatedly,
and never type a password into a non-password field. Use go\_back only to
recover from a wrong page; do not loop between the same two pages.
\item Do not output pixel coordinates, bounding boxes, candidate ids, DOM
details, CSS, or selectors in the "content" field.
\item Choose exactly ONE action per step, and do not state uncertainty
(avoid "maybe", "probably", "I think") in "think".
\end{itemize}
{\bfseries guide\_text RULES:}
\begin{itemize}\itemsep1pt\parskip0pt\topsep1pt
\item guide\_text speaks to the human user in plain language ("Click...",
"Type ... into the search box", "Open the Reports menu", "Scroll down
to..."), and refers to the target by its visible on-screen label or
location.
\item guide\_text must NOT contain candidate ids, numbers from the marks,
coordinates, bounding boxes, DOM, CSS, or selector details.
\item guide\_text must NOT be your reasoning or a prediction of what will
happen (that belongs in "think"); write it as one short direct instruction.
\end{itemize}
Return exactly one JSON object with the five keys above and nothing else.
\end{multicols}
\end{minipage}}}
\end{minipage}}}
\caption{The locked system prompt of the SoM variant, verbatim. The
coordinate variant changes exactly two things: the observation description
tells the model to use the menu to locate the element and act on its on
screen pixel location, and the grounding key is a coordinate pair instead
of \texttt{selected\_candidate\_id}. Every other rule is shared.}
\label{fig:sysprompt}
\end{figure*}

\begin{figure*}[t]
\centering
{\setlength{\fboxsep}{0pt}\setlength{\fboxrule}{0.9pt}%
\fcolorbox[gray]{0.45}{1}{\begin{minipage}{0.96\textwidth}
{\setlength{\fboxsep}{5pt}%
\colorbox[gray]{0.32}{\makebox[\dimexpr\linewidth-10pt][l]{%
\textcolor{white}{\bfseries\small User Message (one step of the worked example; menu abridged)}}}}\par\nointerlineskip
{\setlength{\fboxsep}{8pt}%
\colorbox[gray]{0.97}{\begin{minipage}{\dimexpr\linewidth-16pt}
\footnotesize\ttfamily\raggedright
Task:\\
What is the top-1 best-selling product in 2022\\[4pt]
Action history (what has been done so far; oldest first, empty if this is
the first step):\\
1. Log in with the provided admin account.\\[4pt]
Candidate elements on the current page (each line is "[id] TYPE visible
text"):\\
{}[1] IMG Magento Admin Panel\\
{}[2] A Dashboard\\
{}[3] A Sales\\
{}[5] A Catalog\\
{}[8] A Customers\\
...\\
{}[27] A Sales\\
{}[29] A Reports\\
{}[31] A Stores\\
... \textrm{\textit{(67 lines in the full step)}}\\[4pt]
The current page screenshot, with numbered boxes on the candidate elements
matching the ids above, is attached as an image.\\[4pt]
Decide the single next action and write its guide\_text. Return ONLY one
JSON object with exactly the keys: think, action\_type,
selected\_candidate\_id, content, guide\_text.
\end{minipage}}}\par\nointerlineskip
{\setlength{\fboxsep}{5pt}%
\colorbox[gray]{0.32}{\makebox[\dimexpr\linewidth-10pt][l]{%
\textcolor{white}{\bfseries\small Model Output}}}}\par\nointerlineskip
{\setlength{\fboxsep}{8pt}%
\colorbox[gray]{0.97}{\begin{minipage}{\dimexpr\linewidth-16pt}
\footnotesize\ttfamily\raggedright
\{"think": "Best sellers by year are under the sales reports, so I open
Reports in the left sidebar.",\\
\hphantom{\{}"action\_type": "click",\\
\hphantom{\{}"selected\_candidate\_id": "29",\\
\hphantom{\{}"content": null,\\
\hphantom{\{}"guide\_text": "Open the Reports section from the left
sidebar."\}
\end{minipage}}}
\end{minipage}}}
\caption{The user template filled with one real step of the worked example
task, and a well formed completion for that step.}
\label{fig:usermsg}
\end{figure*}

\section{Metric Definitions and Evaluation Protocol}
\label{app:metrics}

\paragraph{Output contract.}
At every step the model must emit exactly one JSON object with five keys:
\texttt{think}, \texttt{action\_type}, \texttt{selected\_candidate\_id}
(replaced by a coordinate pair under coordinate grounding),
\texttt{content}, and \texttt{guide\_text}. A single locked parser is shared
by data construction, training, and evaluation.

\paragraph{OFCR.}
A step passes the format gate if three conditions hold. (i) Parse: the
output parses as the contract above with nonempty \texttt{think},
\texttt{guide\_text}, and a recognized verb. (ii) Valid action: the verb is
one of the seven in $\mathcal{A}$; the element verbs \texttt{click},
\texttt{type}, and \texttt{select} carry a grounding argument, and a SoM id
must appear in the current candidate menu; \texttt{type} carries text and
\texttt{select} an option label. (iii) No leakage: the guide matches none of
a fixed regular expression family covering mark references (\emph{element
17}, \emph{candidate 17}, \emph{[17]}) and coordinate pairs
(\emph{(312,\,174)}); the full family ships with the harness, and the human
pass of Section~\ref{sec:dataset} backstops paraphrased leakage. OFCR is
the fraction of steps passing all three.

\paragraph{Guide quality and GGS.}
$G_i$ (Equation~\ref{eq:guideq}) joins the predicted guides of task $i$ in
step order and scores them against the joined gold guides with BLEU-1,
BLEU-2, ROUGE-1 (unigram token F1), and ROUGE-L, averaged with equal
weights. Joining is deliberate: no step alignment exists when the predicted
and gold trajectories differ in length, so a step count mismatch simply
lowers overlap; the BLEU brevity penalty and F1 based ROUGE bound length
gaming; and step level fidelity is measured separately by the teacher
forced diagnostics below. Guides that fail OFCR are not additionally zeroed
in $G_i$; the two metrics are reported side by side. A successful run that
takes a valid alternative path is penalized by the single reference, so
part of the guide tax reflects path divergence rather than guide quality;
Section~\ref{sec:coupling} probes this with a guide and success coupling
analysis. Corpus GGS averages Equation~\ref{eq:ggs} over all 174 test
tasks; $\gamma=0.4$, $\tau=120$\,px, the content thresholds, and the Suff
floor were all fixed before any experiment and never tuned.

\paragraph{Teacher forced step diagnostics.}
Given a gold step with candidate set $C$, action correctness decomposes as
$A = \mathrm{Type}\cdot\mathrm{Target}\cdot\mathrm{Content}$.
$\mathrm{Type}$ is exact verb equality. $\mathrm{Target}$ is exact id
equality under SoM; under coordinates it is $1$ if the predicted point falls
inside the gold element's box and $e^{-d/\tau}$ otherwise, where $d$ is the
distance to the box edge and $\tau=120$\,px. $\mathrm{Content}$ compares
payloads by token F1 with thresholds $0.9$ for \texttt{type} and
\texttt{select} and $0.5$ for the \texttt{finish} answer, and is $1$ for the
remaining verbs. Guide faithfulness and sufficiency are
\begin{align}
\mathrm{Faith} &= \tfrac{1}{2}\,\mathrm{Verb} + \tfrac{1}{2}\,\mathrm{Obj},\\
\mathrm{Suff} &=
\begin{cases}
0 & \text{if } m < 0.10,\\
m \cdot p_{\mathrm{tie}} & \text{otherwise,}
\end{cases}
\end{align}
with $m = \max_{c \in C} \mathrm{F1}(\mathrm{anchor}(g),\, v_c)$ and
$p_{\mathrm{tie}} = 0.5$ when the top two candidates lie within $0.10$ of
each other and $1$ otherwise. $\mathrm{Verb}$ checks the guide's head verb
against the action verb through a small lexicon. $\mathrm{Obj}$ is the
token F1 between the guide's quoted anchor span, extracted by
$\mathrm{anchor}(g)$ with fallback to the full sentence, and the step's
target string: the element's visible text for \texttt{click} and
\texttt{select}, the typed text for \texttt{type}, the direction for
\texttt{scroll}, and the answer for \texttt{finish}. $\mathrm{Suff}$ asks
whether the guide identifies a unique referent on the page: a best match
below $0.10$ names nothing, and a near tie makes the reference ambiguous;
whether that referent is the intended one is carried by $\mathrm{Obj}$.
For verbs that target no element, $\mathrm{Suff}$ is $1$ with a clear head
verb and $0.5$ otherwise. The fused step score and
its aggregate are
\begin{equation}
\mathrm{GACS} = F \cdot A \cdot
\sqrt{\mathrm{Faith}\cdot\mathrm{Suff}},
\qquad
\mathrm{SAA} = \overline{A},
\end{equation}
with $F$ the step's OFCR bit. All quantities are deterministic and
dependency free, so diagnostic scoring is identical wherever it runs.

\paragraph{Evaluation protocol.}
Algorithm~\ref{alg:protocol} gives the live protocol, identical under both
grounding schemes; only the action's grounding argument differs. Every
model output consumes one step of the budget and is scored by OFCR; an
action that names no valid target executes nothing on the page. Locally
served models decode greedily; the API model exposes no deterministic mode
and is called with fixed settings (low reasoning effort, 1,024 output
tokens). The LLM judge (GPT-5.5, fixed prompt) scores only the 29 test
tasks whose checker requires semantic comparison of the final answer; it
receives the answer and the reference, never the trajectory or the system
identity, and the same judge scores every run.

\begin{algorithm}[t]
\caption{Live evaluation, one model, one grounding scheme}
\label{alg:protocol}
\begin{algorithmic}[1]
\State reset all site containers to initial snapshots
\State re-register accounts; verify login on every site
\For{each test task $(q, s_0, \Phi, g^{\star})$}
    \State $s \gets s_0$;\quad $h \gets [\,]$ \Comment{guide history}
    \For{$t = 1 \dots 25$}
        \State $x_t \gets$ $1440\times900$ SoM screenshot of $s$
        \State $C_t \gets$ candidate menu from the marks
        \State $(a_t, g_t) \gets \pi_\theta(q, x_t, C_t, h)$ \Comment{fixed decoding}
        \State score $(a_t, g_t)$ with OFCR
        \State execute $a_t$ on the live page; append $g_t$ to $h$
        \If{$\alpha_t = \texttt{finish}$} \textbf{break} \EndIf
    \EndFor
    \State $T \gets$ index of the last executed step
    \State $S \gets \Phi(s_{T+1}, \omega_T)$ \Comment{functional checkers}
    \If{$\Phi$ needs semantic comparison} \State $S \gets$ LLM judge \EndIf
    \State $G_i \gets$ Eq.~\ref{eq:guideq} on $(h, g^{\star})$;\quad
           $\mathrm{GGS}_i \gets$ Eq.~\ref{eq:ggs}
    \State record $S$, OFCR, $G_i$, $\mathrm{GGS}_i$
\EndFor
\end{algorithmic}
\end{algorithm}

\section{Training Details}
\label{app:training}

\begin{table}[H]
\centering
\small
\begin{tabular}{ll}
\toprule
\multicolumn{2}{l}{\textit{Stage 1: SFT (one run per variant)}} \\
\midrule
finetuning        & full parameter, ZeRO-3, bf16 \\
epochs            & 3 \\
learning rate     & $10^{-5}$, cosine schedule \\
batch             & 1 per device $\times$ 8 accumulation \\
loss              & CE on assistant JSON tokens \\
\midrule
\multicolumn{2}{l}{\textit{Stage 2: expert augmented GRPO}} \\
\midrule
initialization    & the SFT variant \\
group             & 6 policy + up to 2 expert rollouts \\
rollout           & temperature 1.0, 25 step budget \\
reward            & binary task success \\
advantage         & mean centered, no std division; \\
                  & zero variance groups dropped \\
clip              & $\varepsilon_l = 0.2$, $\varepsilon_h = 0.28$ \\
KL                & none \\
updates per round & 1 inner epoch, 16 row minibatches \\
learning rate     & $10^{-6}$, gradient clip 0.1 \\
rounds            & 10 (evaluation at 5 and 10) \\
schedule          & expert covered tasks first \\
\bottomrule
\end{tabular}
\caption{Training configuration of both stages.}
\label{tab:training}
\end{table}

Table~\ref{tab:training} lists both configurations. The expert cache is
built before training: GPT-5.5 runs twice over the 407 training tasks
through the harness, at the same budgets as the policy; the two passes
solve 112 and 122 tasks respectively, 139 in union and 95 in both, and only
evaluator verified successes are cached. Each round of GRPO follows
Algorithm~\ref{alg:grpo}; the environments are reset and re authenticated
at every round boundary, and rollouts for one group run in parallel across
six replicated environment sets. All SFT, GRPO training, and live rollouts
ran on a single node with eight NVIDIA RTX PRO 6000 GPUs, using DeepSpeed
for the weight updates and vLLM for the rollouts.

\begin{algorithm}[t]
\caption{Expert augmented GRPO, one round}
\label{alg:grpo}
\begin{algorithmic}[1]
\State reset site containers; re register and verify accounts
\State draw the next scheduled batch of training tasks
\For{each task $q$ in the batch}
    \State roll $\tau_1 \dots \tau_6 \sim \pi_{\theta_{\mathrm{old}}}$
           through the harness \Comment{$T{=}1.0$}
    \If{$q$ is expert covered}
        \State add the cached expert trajectories $\mathcal{E}_q$
    \EndIf
\EndFor
\State merge deferred LLM judge verdicts; $R_i \gets S(\tau_i)$
\State drop zero variance groups; $A_i \gets R_i - \bar{R}$ \Comment{Eq.~\ref{eq:adv}}
\State recompute round start log probabilities with the training stack
\For{each 16 row minibatch}
    \State ascend $J(\theta)$ of Eq.~\ref{eq:grpoobj} \Comment{lr $10^{-6}$}
\EndFor
\State push the updated weights to the inference engine
\end{algorithmic}
\end{algorithm}

\section{Full Results}
\label{app:fullresults}

\paragraph{Run provenance.}
All rows of Table~\ref{tab:main} are single greedy sweeps over the 174
test tasks (the API rows under fixed provider settings,
Appendix~\ref{app:metrics}), scored from saved per step records. OFCR is
recomputed for every row from the raw model outputs with the three tier
definition; the qwen rows decode the raw completions with the locked
parser and recover the candidate menus from the stored prompts. The API
rows and the 9B rows run on two physically distinct replicas of the same
site snapshots with identical harness code, prompts, budgets, and judge;
cross panel comparisons span the two deployments.

\paragraph{Intent template overlap.}
WebArena instantiates tasks from intent templates, and the task level
split leaves templates shared: 159 of the 174 test tasks share a template
with at least one training task; 15 do not. On the 15 unseen template
tasks the API models hold their rates (GPT-5.5 with coordinates solves 7
of 15 against 36.5\% on seen; they receive no training), while the tuned
models' gains concentrate on seen templates: GRPO round 10 SoM reaches
13.8\% on seen template tasks against 1 of 15 on unseen, the same count
as its SFT anchor. The unseen slice is too small for firm conclusions,
but the training track should be read as largely within template
generalization.

\paragraph{Judge agreement.}
The 29 semantic comparison tasks are graded by a GPT-5.5 judge, which also
has systems of its own family under evaluation, so we re graded every saved
final answer of all 15 runs with two independent judges under the identical
grading prompt. Gemini 3.5 Flash agrees with the paper's judge on 94.9\% of
the 435 verdicts and shifts no run by more than two tasks (1.1 points).
Claude Sonnet 4.6 is uniformly more lenient toward every system including
its competitors (agreement 79.1\% with GPT-5.5 and 82.8\% with Gemini),
which raises absolute rates but changes no ordering the paper interprets.
The GPT-5.5 judge shows no self preference: on the GPT-5.5 rows its
verdicts match the independent Gemini judge within one task.

\begin{table}[H]
\centering
\small
\resizebox{\columnwidth}{!}{%
\begin{tabular}{lrrc}
\toprule
Comparison & $\Delta$SR & a/b & 95\% CI \\
\midrule
SoM r10 vs SFT           & $+6.3$  & 17/6  & $[+1.1, +11.5]$ \\
SoM r5 vs SFT            & $+4.0$  & 13/6  & $[-1.1, +9.2]$ \\
SoM r10 vs r5            & $+2.3$  & 8/4   & $[-1.7, +6.3]$ \\
coord r5 vs SFT          & $+1.7$  & 10/7  & $[-2.9, +6.3]$ \\
coord r10 vs SFT         & $+0.6$  & 7/6   & $[-3.4, +4.6]$ \\
Gemini SoM vs coord      & $+13.8$ & 34/10 & $[+6.9, +21.3]$ \\
GPT-5.5 coord vs SoM     & $+1.7$  & 14/11 & $[-4.0, +7.5]$ \\
Claude SoM vs coord      & $+0.6$  & 13/12 & $[-5.2, +6.3]$ \\
GPT-5.5 coord vs SoM r10 & $+24.1$ & 47/5  & $[+17.2, +31.6]$ \\
\bottomrule
\end{tabular}}
\caption{Paired comparisons on the 174 test tasks: $\Delta$SR in points,
tasks solved only by the first / only by the second system, and task level
bootstrap 95\% CIs (10k resamples, fixed seed). The difference between the
SoM and coordinate training gains is $+5.7$ points with CI
$[-0.6, +12.1]$.}
\label{tab:significance}
\end{table}

\begin{table}[H]
\centering
\small
\begin{tabular}{lrrrr}
\toprule
Gold length & 2--5 & 6--8 & 9--12 & 13+ \\
$n$ tasks   & 71   & 44   & 15    & 37  \\
\midrule
SFT SoM      & 11.3 & 2.3  & 0.0  & 2.7 \\
GRPO r5      & 18.3 & 6.8  & 0.0  & 2.7 \\
GRPO r10     & \textbf{21.1} & \textbf{11.4} & 0.0 & 2.7 \\
GPT-5.5 SoM  & 32.4 & 40.9 & 20.0 & 37.8 \\
GPT-5.5 coord& \textbf{39.4} & \textbf{43.2} & 20.0 & 35.1 \\
Gemini SoM   & 33.8 & 38.6 & 20.0 & 29.7 \\
Claude SoM   & 23.9 & 31.8 & 20.0 & 32.4 \\
\bottomrule
\end{tabular}
\caption{Success rate (\%) by gold demonstration length, over the 167 test
tasks with annotated trajectories. GRPO gains concentrate on short and
medium tasks; every 9B variant collapses beyond eight gold steps, where the
API models sustain 20 to 38\%.}
\label{tab:strata}
\end{table}

\paragraph{Solved set overlap.}
GRPO round 10 (SoM) solves 17 tasks its SFT initialization could not and
loses 6; from round 5 to 10 it keeps 15 of 19, loses 4, and gains 8, so
acquisition continues through the self rollout phase. Its SoM and
coordinate successes overlap on only 8 tasks; the union covers 29 of 174
(16.7\%), an immediate headroom for modality routing.

\paragraph{Action distribution.}
Over all steps of the SoM runs: the base model emits 81.4\% unparseable
steps (its OFCR); SFT eliminates these (2.3\%) but scrolls on 34.9\% of
steps; GRPO round 10 shifts mass to decisive interaction (click 57.9\%,
scroll 13.5\%, type 16.2\%) and rediscovers go\_back (2.3\%). The GPT-5.5
SoM run distributes 69.5\% click, 16.0 type, 5.2 scroll, 4.1 finish, 2.5
press\_enter, 1.5 select, and 1.2 go\_back.

\section{pass@6 Details}
\label{app:pass6}

\begin{figure}[H]
\centering
\includegraphics[width=0.95\columnwidth]{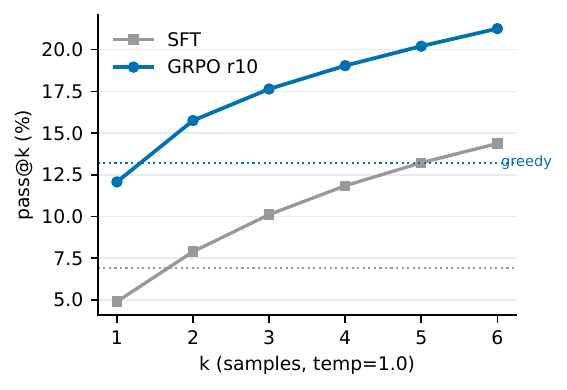}
\caption{pass@$k$ under six independent samples per task at temperature
1.0 (dotted lines: greedy SR). A pure sharpening account of GRPO would
predict a flat gap at $k=6$.}
\label{fig:passk}
\end{figure}

Six trajectories per task are sampled at temperature 1.0, one per
environment set (1{,}044 episodes per model), same harness and judge as
the main runs. SFT SoM: greedy 6.9\%, sampled pass@1 4.9\%, pass@3 10.1\%,
pass@6 14.4\%, 149 never solved tasks. GRPO round 10: greedy 13.2\%,
pass@1 12.1\%, pass@3 17.6\%, pass@6 21.3\%, 137 never solved. The paired
pass@6 gap is $+6.9$ points (21 tasks only GRPO, 9 only SFT; task
bootstrap 95\% CI $[+1.2, +13.2]$). Figure~\ref{fig:passk}
shows the curves: GRPO lifts both the single sample rate and the
exploration ceiling, and its sampled pass@1 nearly matches its greedy rate,
while SFT loses success under sampling.

\section{Plain GRPO Runs without Expert Injection}
\label{app:failedruns}

\begin{figure}[H]
\centering
\includegraphics[width=0.95\columnwidth]{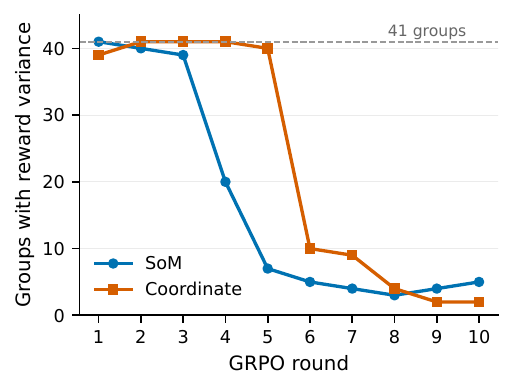}
\caption{Groups with reward variance (out of 41 per round) in the expert
augmented runs. While the schedule serves expert covered tasks (rounds 1
to 3 for SoM, 1 to 5 for coordinates), nearly every group carries a
gradient; once coverage is exhausted, variance collapses to the few tasks
the policy can already sometimes solve.}
\label{fig:keptgroups}
\end{figure}

Before expert injection we ran ten plain GRPO configurations (G = 6 self
rollouts, no expert rows) spanning shaped and binary rewards, curricula,
premature finish penalties, and KL variants, for 1 to 13 rounds each.
Across the nine runs with at least two logged rounds, the first to last
round training batch success deltas were $-9.4$, $-7.3$, $-10.4$, $-5.2$,
$+24.0^{*}$, $+1.0$, $-10.4$, $-8.3$, and $-14.6$ points; the starred
outlier used an inflated stub judge during a period without API egress,
and its clean re evaluation measured 6\%. Two caveats make this
documentary rather than a controlled ablation: the numbers are training
batch statistics on freshly sampled tasks (not the fixed test set), and
most of these runs predate the deterministic judge routing. The mechanism
is nonetheless visible in the expert augmented runs themselves
(Figure~\ref{fig:keptgroups}): reward variance, the precondition for any
GRPO gradient, tracks expert coverage almost exactly.

\section{Qualitative Case Studies}
\label{app:cases}

\begin{figure*}[t]
\centering
\includegraphics{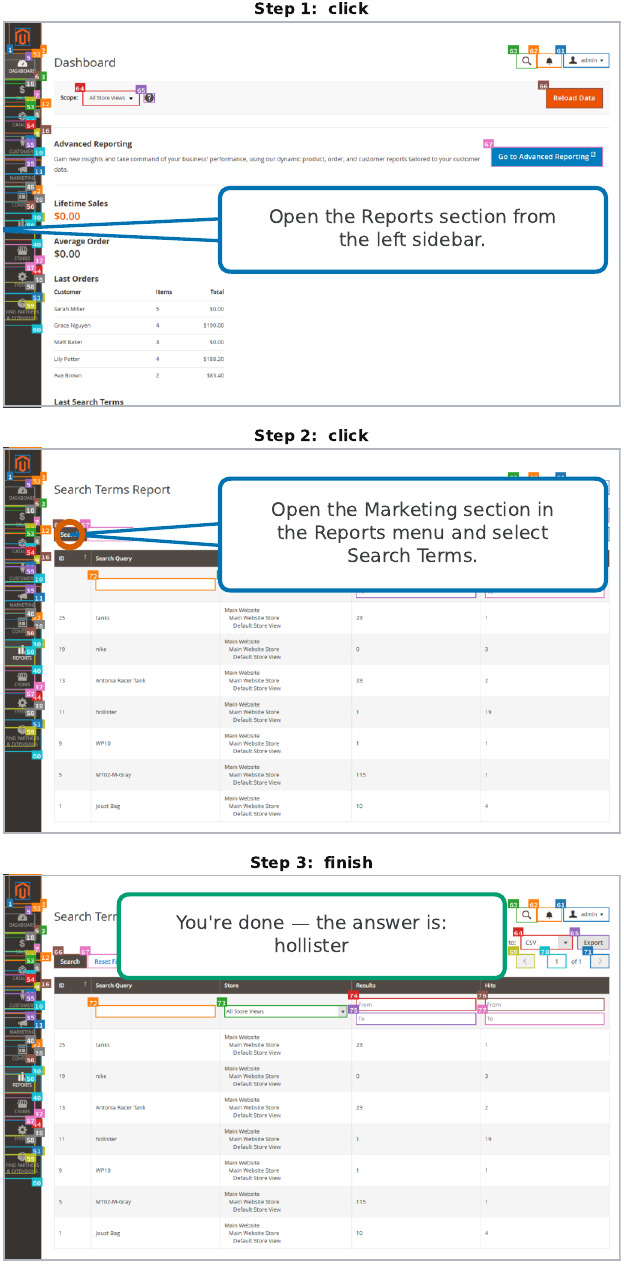}
\caption{Gold trajectory for the shopping admin task \emph{List the top 1
search terms in my store}: open Reports, open the Search Terms report,
finish with the answer read from the result table.}
\label{fig:case_admin}
\end{figure*}

\begin{figure*}[t]
\centering
\includegraphics{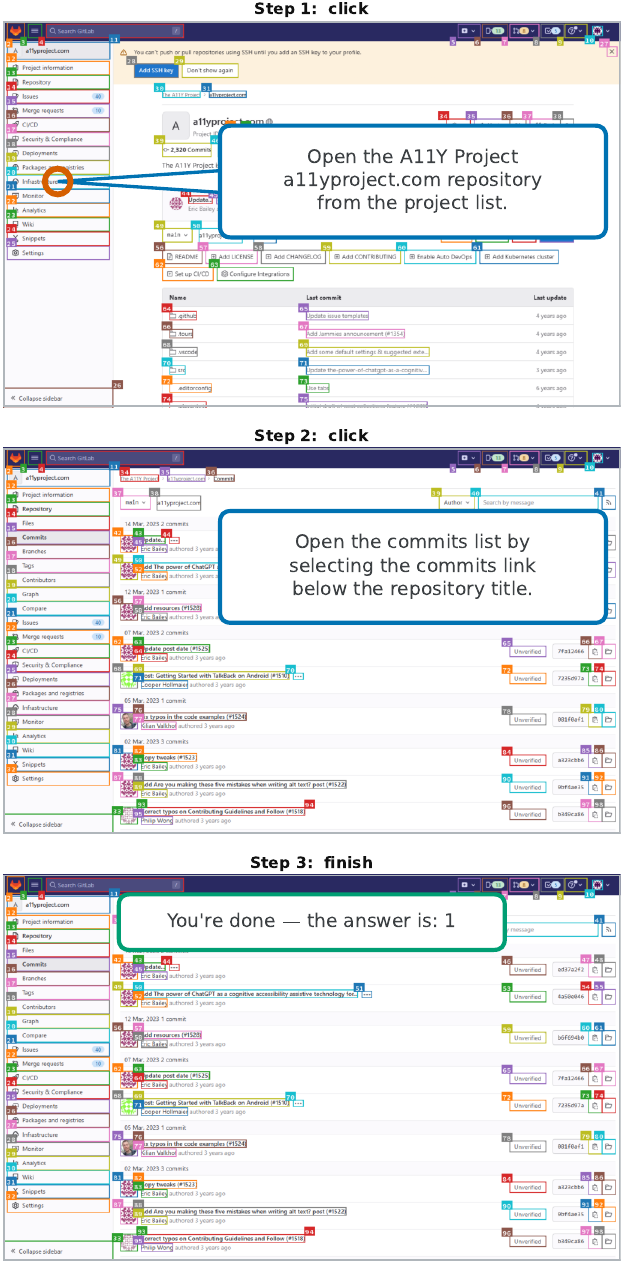}
\caption{Gold trajectory for the GitLab task \emph{How many commits did
kilian make to a11yproject on 3/5/2023}: open the repository, open its
commit history, finish with the count.}
\label{fig:case_gitlab}
\end{figure*}

\begin{figure*}[t]
\centering
\includegraphics[width=0.98\textwidth]{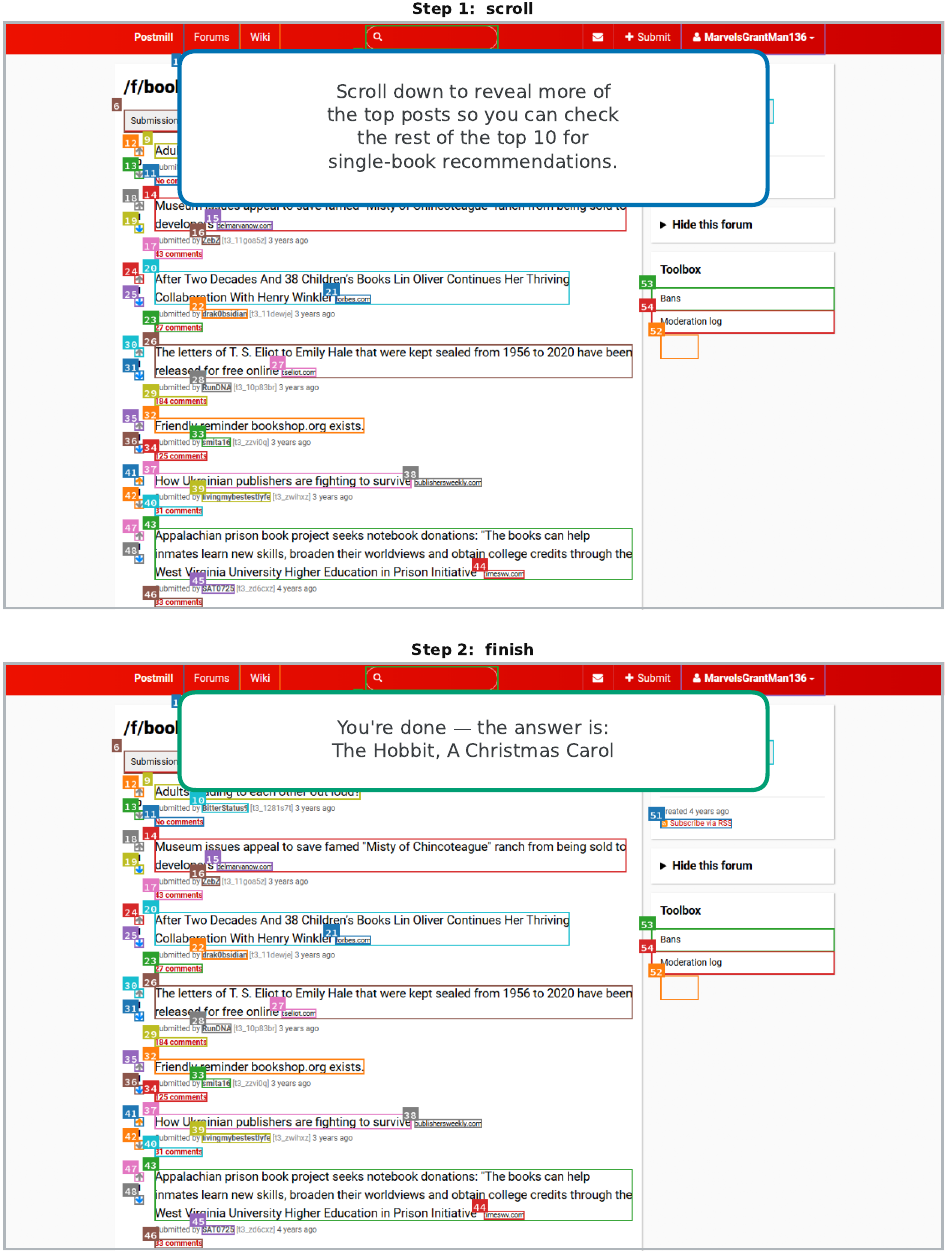}
\caption{Gold trajectory for the Reddit task asking for single book
recommendations among the top posts of the books forum: scroll to reveal
the remaining posts, then finish with the two titles.}
\label{fig:case_reddit}
\end{figure*}

\begin{figure*}[p]
\centering
\includegraphics{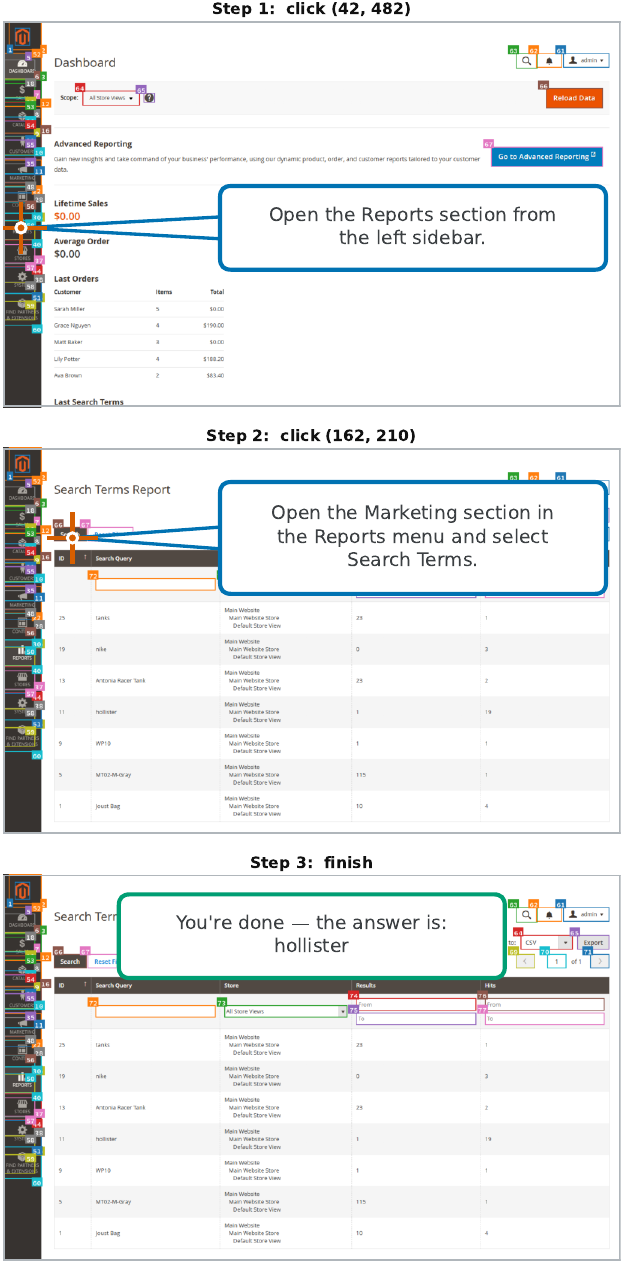}
\caption{Coordinate view of the shopping admin task of
Figure~\ref{fig:case_admin}: same pages and guides, actions grounded as
pixel positions.}
\label{fig:case_coord_admin}
\end{figure*}

\begin{figure*}[p]
\centering
\includegraphics{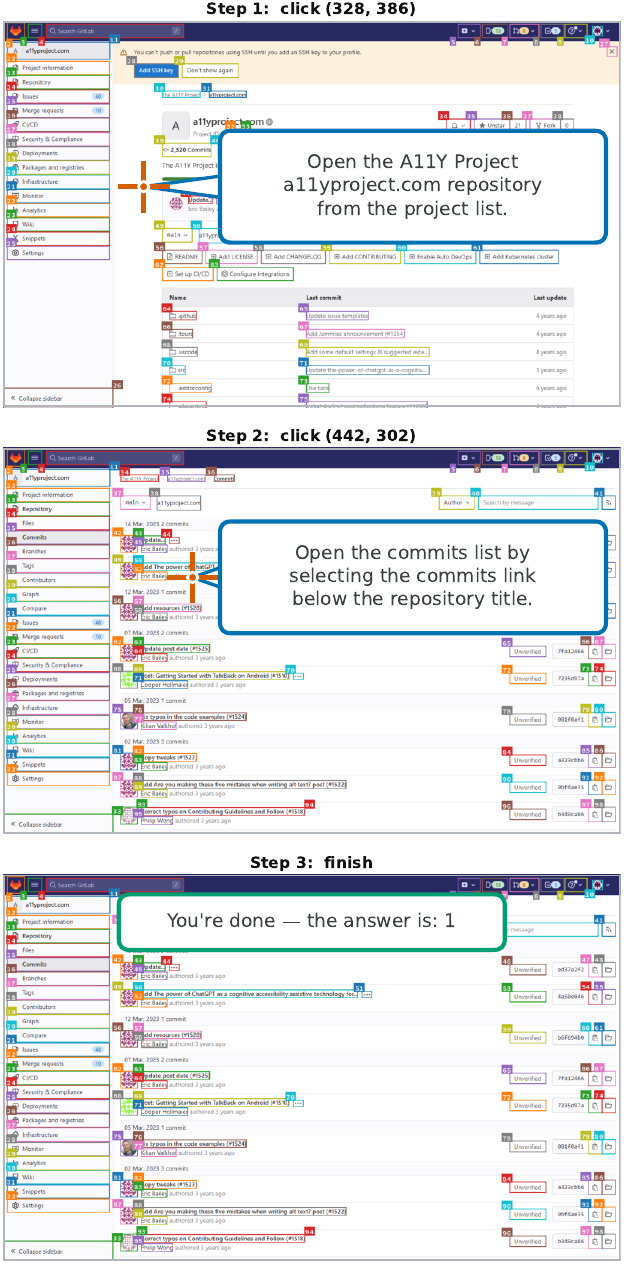}
\caption{Coordinate view of the GitLab task of
Figure~\ref{fig:case_gitlab}.}
\label{fig:case_coord_gitlab}
\end{figure*}

\begin{figure*}[p]
\centering
\includegraphics[width=0.98\textwidth]{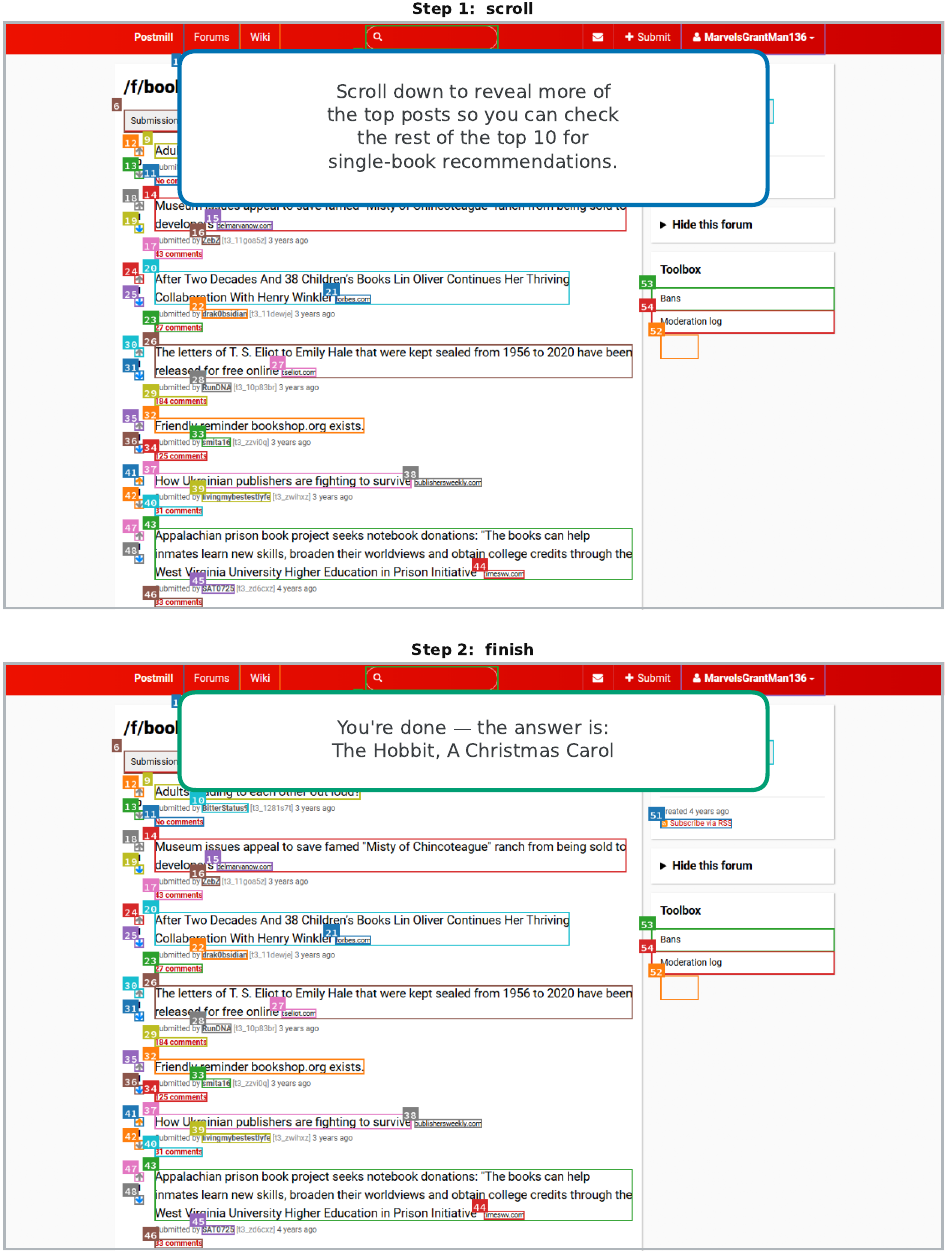}
\caption{Coordinate view of the Reddit task of
Figure~\ref{fig:case_reddit}.}
\label{fig:case_coord_reddit}
\end{figure*}

Figures~\ref{fig:case_admin}, \ref{fig:case_gitlab}, and \ref{fig:case_reddit}
render three complete gold trajectories the way a user would experience
them: at every step, the guide sentence appears in a callout anchored on
the target element of the real page, and the final step announces the
answer. The three tasks are among the shortest in the corpus and cover three sites and three verbs (click, scroll, finish). Figures~\ref{fig:case_coord_admin}, \ref{fig:case_coord_gitlab}, and \ref{fig:case_coord_reddit} show the coordinate view of the same three tasks: identical screenshots and gold guides, with each action grounded as a raw pixel position (crosshair, coordinates in the step header) instead of a mark id, the shared observation design of Section~\ref{sec:task}. Together they illustrate the intended product shape of MAG from Figure~\ref{fig:teaser}: the agent's two outputs per step, an action and a guide, are exactly the assets an in app guidance overlay needs.

\end{document}